\documentclass[11pt]{article}
\usepackage[margin=1in]{geometry}
\usepackage{amsmath,amssymb,amsthm}
\usepackage{algorithm}
\usepackage{algpseudocode}
\usepackage{graphicx}
\usepackage{thmtools}
\usepackage{hyperref}
\usepackage{cleveref}
\usepackage{xcolor}

\newtheorem{definition}{Definition}
\newtheorem{theorem}{Theorem}
\newtheorem{lemma}{Lemma}
\newtheorem{corollary}{Corollary}

\theoremstyle{remark}

\newcommand{\remove}[1]{}

\title{Nearly Linear Private Learning of Littlestone Classes}
\author{%
  Shay Moran\thanks{Departments of Mathematics,
  Computer Science, and Data and Decision Sciences,
  Technion -- Israel Institute of Technology;
  Google Research. Shay Moran acknowledges support by Israel by ISF grant 2362/26. and by the European Union (ERC, GENERALIZATION, 101039692).}
  \and
  Chao Yan\thanks{Department of Mathematics,
  Technion -- Israel Institute of Technology. Chao Yan acknowledges support by Israel by ISF grant 2362/26. and by the European Union (ERC, GENERALIZATION, 101039692).}
}
\date{}
\begin{document}
\maketitle

\begin{abstract}
Differentially private learning aims to learn from data while protecting
individual training examples. For binary classification, finite
Littlestone dimension characterizes approximate private PAC learnability
(Alon et al., JACM 2022). Understanding how this dimension
controls private sample complexity remains a central question. For classes of Littlestone dimension $d$, the first general upper bound
was doubly exponential in $d$ (Bun, Livni, and Moran,
FOCS 2020). Subsequent work improved this to
$\widetilde O(d^6)$ (Ghazi et al., STOC 2021)
and then to $\widetilde O(d^5)$ (Lyu, STOC 2026).
We prove a nearly linear upper bound of $O(d\log d)$.

Our main algorithmic contribution is a structural exponential mechanism
for data-dependent candidate sets equipped with a rank satisfying
suitable stability conditions. For natural Littlestone classes such as
subspace indicators, this rank has a simple geometric interpretation:
it is the codimension. This yields a private PAC learner for
subspaces of dimension at most $d$ over any field, with tight linear
dependence on $d$ and no dependence on the ambient dimension, improving
the quadratic dependence of Kaplan et al. (STOC 2025).
Our second contribution is combinatorial: drawing on model-theoretic
ideas, we construct a suitable rank for general Littlestone classes
and establish the counting bounds required by the mechanism.
\end{abstract}

\section{Introduction}
\label{sec:introduction}

A basic goal of learning theory is to understand how much data is
needed to learn an unknown concept. In the classical PAC model, a
learner receives labeled examples drawn from an unknown distribution
and should produce a predictor that is accurate on new examples from
the same distribution; see, for example,~\cite{Kearns94}.
For binary concept classes, Vapnik--Chervonenkis (VC)
dimension~\cite{VC} characterizes learnability and controls the number
of examples needed to achieve a desired accuracy and
confidence~\cite{BEHW,EHKV,Talagrand94,Hanneke16}.

When training examples contain information about individuals, we
also want the learning algorithm to protect their privacy.
Differential privacy formalizes the protection of individuals by
requiring that the distribution of the algorithm's output change
only slightly when a single training example changes~\cite{DiNi03,DMNS06}.
There are two standard forms of this guarantee: \emph{pure}
differential privacy and its relaxation, \emph{approximate
differential privacy}~\cite{DKMMN06}. Every purely private
algorithm is also approximately private, but the converse need
not hold.
\begin{samepage}
Private PAC learning combines privacy with the usual prediction
guarantee~\cite{KLNRS08}. This raises three basic questions:
\begin{enumerate}
\setlength{\itemsep}{0pt}
\item \textbf{Learnability:} Which concept classes can be learned privately?
\item \textbf{Sample complexity:} How many examples are necessary and sufficient?
\item \textbf{Learning principles:} Which general learning methods attain the optimal sample complexity?
\end{enumerate}
\end{samepage}

For pure differential privacy, all three questions have a general
answer. Finiteness of the \emph{representation dimension}
characterizes which concept classes are learnable, and this
dimension determines their optimal sample complexity up to
constant factors for fixed learning and privacy
parameters~\cite{BNS13}.
The matching upper bound is achieved by the \emph{exponential
mechanism}~\cite{MT07}, a basic scheme for private optimization.
It is thus the governing algorithmic principle behind this
characterization of pure private PAC learning.
We recall the mechanism in Section~\ref{sec:structural-em}.

For approximate differential privacy, the first question has a
complete answer: a binary concept class is privately PAC learnable
if and only if its Littlestone dimension is finite
\cite{AlonLMM19,bun2020equivalence,ALBMM22}.
Finite Littlestone dimension also characterizes online
learnability~\cite{littlestone87}. Thus, exactly the same classes
are learnable when examples arrive stochastically, under approximate
differential privacy, and when examples arrive adversarially,
without a privacy requirement.

The other two questions---the optimal sample complexity and the
algorithmic principles that attain it---remain open.
We now review what is known about the sample complexity of private learning. Let $d$ and $v$ denote the Littlestone dimension and VC dimension of the concept class, respectively.
Suppressing the learning and privacy parameters, the general
upper bound improved from doubly exponential in
$d$~\cite{bun2020equivalence} to
$\widetilde O(d^6)$~\cite{ghazi2021sample} and then to
$\widetilde O(d^5)$~\cite{lyu2026private}, while the general
lower bound is $\Omega(v+\log^*d)$~\cite{EHKV,AlonLMM19}.

We prove a nearly linear upper bound of $O(d\log d)$ (Theorem~\ref{thm:private-littlestone-pac}).
Among bounds depending only on Littlestone dimension, this is
optimal up to a logarithmic factor. Indeed, even without privacy,
learning requires $\Omega(v)$ examples, where $v$ is the VC
dimension~\cite{EHKV}. For classes with $v=d$, this gives an
$\Omega(d)$ lower bound.

Our proof has two main ingredients. First, we formulate a
\emph{structural exponential mechanism}, a variant of the
exponential mechanism for data-dependent candidate sets (Theorem~\ref{thm:structural-em}).
To apply this variant, one must equip the candidate hypotheses
with a ranking function satisfying suitable stability axioms.
Second, we construct such a rank for general Littlestone classes (Definition~\ref{def:rank})
and verify these axioms (Section~\ref{sec:littlestone-axioms}). 
This rank is rooted in model theory and builds on connections with learning
developed by~\cite{MalliarisMoran2021,MalliarisMoran2022}.
As a simpler application, we use the structural mechanism to
learn indicators of linear subspaces. Here codimension serves as
the rank, giving a tight sample complexity bound that is linear
in the maximum subspace dimension and independent of the ambient
dimension (Theorem~\ref{thm:private-linear-subspaces}).

\begin{samepage}
\section{Main Results}
\label{sec:main-results}

We use standard notions from learning theory and differential privacy,
including PAC learning, VC and Littlestone dimensions, and pure and
approximate differential privacy. Readers unfamiliar with these notions
may consult the technical preliminaries in
Section~\ref{sec:preliminaries}.
We use $\alpha$ for the accuracy parameter and $\beta$ for the failure
probability, so the learning guarantee holds with confidence $1-\beta$.
The privacy parameters are $\varepsilon$ and $\delta$.
All privacy guarantees use addition/deletion adjacency, as defined
in Section~\ref{sec:sem-main-result}.
Throughout, $0<\alpha,\beta,\delta<1$ and $0<\varepsilon\le1$.
The results concern sample
complexity, without asserting computational efficiency.

The learning guarantees below are stated in the realizable setting.
Known reductions~\cite{AlonBMS20,LWY25} extend such guarantees to the
agnostic setting while preserving properness. The refined reduction
of~\cite{LWY25} incurs an additive sample overhead of
$\widetilde O(\operatorname{VCdim}(\mathcal H)/\alpha^2)$,
where $\alpha$ denotes excess error. The precise bound and its
application under addition/deletion adjacency are given in
Appendix~\ref{sec:agnostic-extension}.

\end{samepage}

\begin{samepage}
\subsection{Nearly Linear Dependence on Littlestone Dimension}

\begin{theorem}[Private PAC learning of Littlestone classes]
\label{thm:private-littlestone-pac}
Let $\mathcal H\subseteq\{0,1\}^{\mathcal X}$ be a nonempty class
with $\operatorname{LDim}(\mathcal H)=d\ge1$.
There exists an $(\varepsilon,\delta)$-differentially private
$(\alpha,\beta)$-PAC learner for $\mathcal H$ with sample complexity
\[
O\!\left(
\frac{d\log\frac{d}{\alpha\varepsilon\delta}
+\log\frac1\beta}{\alpha\varepsilon}
\right).
\]
\end{theorem}
\end{samepage}

The construction and proof appear in
Section~\ref{sec:littlestone-proof}.
For fixed learning and privacy parameters, the bound is $O(d\log d)$.
The lower bound $\Omega(d)$ for classes with
$\operatorname{VCdim}(\mathcal H)=\operatorname{LDim}(\mathcal H)=d$
shows that this dependence is optimal up to a logarithmic factor
in the worst case.

Although our bound is nearly optimal in the worst case, it leaves
room for substantially smaller sample complexity when VC dimension
is much smaller than Littlestone dimension.
Writing $v=\operatorname{VCdim}(\mathcal H)$, the
\emph{log-star question} asks whether sample complexity can be
bounded by $\operatorname{poly}(v,\log^*d)$, apart from the
learning and privacy parameters.
This question, explicitly raised by Nissim, Stemmer, and Tsfadia
in their COLT 2026 open-problem paper~\cite{NST26}, remains open.

\subsection{A Structural Exponential Mechanism}
\label{sec:sem-main-result}
\label{sec:structural-em}

The exponential mechanism~\cite{MT07} is a basic tool for private
optimization and can be viewed as a differentially-private counterpart of
\emph{empirical risk minimization}. In learning, an empirical risk
minimizer selects a hypothesis with the smallest \emph{empirical loss} on the sample. In the terminology of
the exponential mechanism, one instead maximizes a \emph{score};
taking the score to be minus the loss makes the two objectives
equivalent. Such a deterministic choice can change abruptly when one
example changes. The exponential mechanism instead assigns larger
probability to hypotheses with higher scores, while controlling how
much these probabilities change between neighboring datasets.

Throughout, a sample (or dataset) is a finite sequence of examples,
with repetitions allowed. Two datasets are \emph{neighboring} if
one is obtained from the other by inserting one example at any
position or deleting one entry, preserving the relative order of
all remaining entries. All privacy
guarantees use this \emph{addition/deletion adjacency} convention.
Our PAC sample complexity bounds also hold under replacement
adjacency.\footnote{\label{fn:replacement-privacy}Under replacement
adjacency, two sequences have the same length and differ in exactly one coordinate.
A replacement consists of a deletion followed by an addition.
By group privacy, an $(\eta,\tau)$-private algorithm under
addition/deletion is $(2\eta,(1+e^\eta)\tau)$-private under
replacement. Thus, for $\varepsilon\le1$, using
$\eta=\varepsilon/2$ and $\tau=\delta/3$ gives
$(\varepsilon,\delta)$-privacy under replacement, affecting our
sample complexity bounds only by constant factors.}

The exponential mechanism guarantees pure differential privacy ($\delta=0$) and is the governing algorithmic principle
behind the characterization of pure private PAC learning by
representation dimension~\cite{BNS13}. 

\paragraph{The classical exponential mechanism.}
Let $\mathcal R$ be a fixed, finite, nonempty output set, 
and let~$q_S(h)$ be a real-valued score to be maximized.
Suppose that $|q_S(h)-q_T(h)|\le\Delta$ for every pair of neighboring
datasets $S,T$ and every $h\in\mathcal R$, where $\Delta>0$.
The exponential mechanism~\cite{MT07} samples $h\in\mathcal R$
with probability proportional to
$\exp(\varepsilon q_S(h)/(2\Delta))$.
It is $\varepsilon$-differentially private and, for every input $S$
and $0<\beta<1$, with probability at least $1-\beta$,
\[
q_S(\widehat h)\ge
\max_{h\in\mathcal R}q_S(h)
-\frac{2\Delta}{\varepsilon}\log\frac{|\mathcal R|}{\beta}.
\]
See also~\cite[Section~3.4]{DR14}.

\paragraph{A structural variant.}
The classical mechanism requires its candidate set to be fixed
independently of the dataset. This prevents a direct application
when the useful candidates are constructed from the sample itself:
an output available on one dataset may disappear on its neighbor.
Approximate privacy allows such outputs to have a small total
probability, which can be charged to $\delta$.
Our main algorithmic contribution is a structural variant of the
exponential mechanism that controls this probability and permits
data-dependent candidate sets. It underlies our nearly linear
dependence on Littlestone dimension.

The mechanism modifies the empirical score by adding a structural
bonus $\lambda\rho(h)$, where $\rho$ is an integer rank.
Higher rank is favored.
This balance between empirical fit and a preference for simpler
solutions parallels structural risk minimization and motivates the
name \emph{structural exponential mechanism}.

\paragraph{Setup.}
Fix a (possibly infinite) output universe $\mathcal U$ and a rank
$\rho:\mathcal U\to\mathbb Z_{\ge0}$.
For each dataset $S$, let
$q_S:\mathcal U\to\mathbb R$ be a score function and let
$\operatorname{Out}(S)\subseteq\mathcal U$ be a finite, nonempty
\emph{data-dependent output set}.
The score is to be maximized. For example, in classification,
$q_S(h)$ can count the examples that $h$ classifies correctly.
Write $R_S$ for the range of the rank on $\operatorname{Out}(S)$:
\begin{equation}
R_S=\max_{h\in\operatorname{Out}(S)}\rho(h)
-\min_{h\in\operatorname{Out}(S)}\rho(h).
\label{eq:rank-range}
\end{equation}

\begin{algorithm}[H]
\caption{Structural exponential mechanism $\mathcal A$}
\label{alg:structural-em}
\begin{algorithmic}[1]
\Require Dataset $S$; privacy parameters
$0<\varepsilon\le1$, $0<\delta<1$.
\Require Score sensitivity bound $\Delta>0$ and preimage growth bound
$B\ge1$, as in Theorem~\ref{thm:structural-em}.
\State Set $\displaystyle
\lambda\gets\Delta+\frac{2\Delta}{\varepsilon}\log\frac{8B}{\delta}$.
\State Sample $\widehat h\in\operatorname{Out}(S)$ with probability
proportional to
\[
\exp\!\left(\frac{\varepsilon}{2\Delta}
[q_S(\widehat h)+\lambda\rho(\widehat h)]\right).
\]
\State \Return $\widehat h$.
\end{algorithmic}
\end{algorithm}

\begin{theorem}[Structural exponential mechanism]
\label{thm:structural-em}
Fix $\Delta>0$ and $B\ge1$, and suppose that for every ordered pair of neighboring datasets $(S,T)$
there is a map
$F_{S,T}:\operatorname{Out}(S)\to\operatorname{Out}(T)$
satisfying the following conditions:
\begin{enumerate}
\item \textbf{Low sensitivity.}
For every $h\in\mathcal U$, $|q_S(h)-q_T(h)|\le\Delta$.

\item \textbf{Rank increase.}
Let \(g\in\operatorname{Out}(S)\). If \(g\in\operatorname{Out}(T)\) then $F_{S,T}(g)=g$;
else,
\(\rho(F_{S,T}(g)) > \rho(g)\).

\item \textbf{Bounded preimages.}
For every $h\in\operatorname{Out}(T)$ and integer $t\ge1$,
\[
\left|\left\{
g\in F_{S,T}^{-1}(h):\rho(g)=\rho(h)-t
\right\}\right|\le B^t.
\]

\item \textbf{Score preservation under the maps.}
For every $h\in\operatorname{Out}(S)$,
$|q_S(h)-q_T(F_{S,T}(h))|\le\Delta$.
\end{enumerate}
Then, Algorithm~\ref{alg:structural-em} is
$(\varepsilon,\delta)$-differentially private and
for every dataset $S$ and $0<\beta<1$, with probability at
least $1-\beta$ its output satisfies
\begin{equation}
q_S(\widehat h)\ge
\max_{h\in\operatorname{Out}(S)}q_S(h)
-\frac{2\Delta}{\varepsilon}\left[
R_S\log\frac{16B}{\delta}
+\log\frac{|\operatorname{Out}(S)|}{\beta}\right].
\label{eq:structural-em-utility}
\end{equation}
\end{theorem}

The proof appears in Section~\ref{sec:mechanism-proof}.
{Equation~\eqref{eq:structural-em-utility} bounds the difference
between the output's score and the highest score among the
candidates. The bound consists of the usual exponential-mechanism
term, logarithmic in the number of candidates, and an additional
term linear in the rank range $R_S$ defined in~\eqref{eq:rank-range}.
The term $R_S\log B$ reflects the local growth of preimages:
the bound $B^t$ at rank gap $t$ contributes a logarithmic cost
of $\log B$ per unit of rank, over the range $R_S$.
In classification, taking $q_S(h)=|S|-\ell_S(h)$, where
$\ell_S(h)$ is the integer number of mistakes on $S$, gives
$\Delta=1$. In our application to a class $\mathcal H$
of Littlestone dimension $d$, the rank takes values in
$\{0,\ldots,d\}$, so $R_S\le d$; see Section~\ref{sec:littlestone-overview}.
\par}

Note that the maps $F_{S,T}$ are used only in the proof and are not needed for the algorithm. 

\medskip

In the next section we explain how this mechanism recovers the classical
exponential mechanism and applies to subspace learning and general
Littlestone classes.
That section provides an overview; the formal proofs of the mechanism
and its applications appear in Sections~\ref{sec:mechanism-proof},
\ref{sec:subspace-proof}, and~\ref{sec:littlestone-proof}.

\section{Technical Overview}
\label{sec:overview}
This section gives an informal overview of the structural exponential
mechanism and the combinatorial ideas behind its application to
Littlestone classes. We begin with a simple warm-up: recovering the
classical exponential mechanism by assigning rank zero to every
candidate. We then turn to a richer application, learning subspaces,
where codimension provides a natural rank. This yields a tight
sample complexity bound, improving the quadratic dependence on
subspace dimension in the bound of Kaplan et al.~\cite{KMMST25}
to linear. Building on the intuition from subspaces, we explain
how to construct an abstract rank for general Littlestone classes
and establish the counting bounds required by the mechanism.
Formal proofs are deferred to the subsequent sections.

\subsection{The Classical Exponential Mechanism}
We now explain how to recover the classical exponential mechanism
as an instance of the structural exponential mechanism.
To this end, set the output universe $\mathcal U$ to be
$\mathcal R$, the fixed, finite, nonempty set of candidate solutions
available to the classical mechanism, and set
$\operatorname{Out}(S)=\mathcal R$ for every sample $S$.
Thus, the same candidates are available on every sample.

We use the original score function $q_S$, with sensitivity at most
$\Delta$, and assign rank zero to every candidate:
\[
\rho(h)=0\quad\text{for every }h\in\mathcal R.
\]
These choices specify the sampling rule. The rank bonus vanishes,
so Algorithm~\ref{alg:structural-em} samples $h\in\mathcal R$ with
probability proportional to $\exp(\varepsilon q_S(h)/(2\Delta))$,
exactly as in the classical exponential mechanism.

To apply Theorem~\ref{thm:structural-em}, we must also define the
maps used in the analysis. For every ordered pair of neighboring
datasets $(S,T)$, take the identity map
$F_{S,T}:\mathcal R\to\mathcal R$, given by $F_{S,T}(h)=h$.
These maps are used only in the analysis and need not be computed
by the algorithm.
Since all candidates belong to both output sets, the identity map
satisfies the rank condition. It has no preimages of smaller rank,
so the preimage bound holds with $B=1$. Score preservation under
this map is exactly the assumed sensitivity bound. Thus all the
hypotheses of Theorem~\ref{thm:structural-em} are satisfied.
The rank range is $R_S=0$, so the theorem gives the usual utility
bound with no rank term. Moreover, no outputs disappear between
neighboring datasets, so the privacy analysis gives pure
$\varepsilon$-privacy.

\subsection{Subspaces: Codimension as Rank}
\label{sec:subspace-overview}

We next explain how to instantiate the structural exponential
mechanism for private learning of subspace indicators. Here the
candidate set depends on the sample, and codimension provides
a natural rank.

Let $\mathbb F$ be any field and let $1\le d\le D$.
The concept class is
\[
\mathcal H_{d,D}
=\{h_V:x\mapsto\mathbf 1\{x\in V\}:
V\le\mathbb F^D,\ \dim V\le d\}.
\]
Here $V\le\mathbb F^D$ means that $V$ is a linear subspace.
The learner receives independent labeled examples
$(x,h_{V_*}(x))$, where the target subspace $V_*$ has dimension at
most $d$ and the distribution of $x$ is arbitrary.
The goal is small population classification error, counting both
false positives and false negatives.

\begin{theorem}[Private PAC learning of subspaces]
\label{thm:private-linear-subspaces}
For every field $\mathbb F$ and integers $1\le d\le D$, the class
$\mathcal H_{d,D}$ has a proper
$(\varepsilon,\delta)$-differentially private
$(\alpha,\beta)$-PAC learner with sample complexity
\[
O\!\left(
\frac{d\log\frac{1}{\alpha\varepsilon\delta}
+\log\frac1\beta}{\alpha\varepsilon}
\right).
\]
The bound is independent of the field and the ambient dimension $D$.
\end{theorem}

This class has VC dimension and Littlestone dimension exactly $d$
(Lemma~\ref{lem:subspace-dimensions}). Consequently, for fixed
learning and privacy parameters, the linear dependence on $d$ is
tight even without privacy.
Theorem~\ref{thm:private-linear-subspaces} is sharper than what follows
by substituting $\operatorname{LDim}(\mathcal H_{d,D})=d$ into
Theorem~\ref{thm:private-littlestone-pac}: it avoids the additional
$\log d$ factor.

Kaplan et al.~\cite{KMMST25} gave efficient private algorithms for
subspace learning with quadratic dependence on the target dimension.
They explicitly asked about closing the gap between their bounds
and the necessary linear dependence on dimension
\cite[Section~2.4 and Appendix~A]{KMMST25}.
Our bound closes this dimensional gap at the level of sample
complexity. Their open question also asks for computationally
efficient algorithms; our result does not settle that additional
requirement.
Combining our bound with their reduction yields an
$(\varepsilon,\delta)$-differentially private solver for linear
equations over any field. Given a consistent system in $d$
variables with sufficiently many equations, it returns an
assignment satisfying at least a $(1-\alpha)$ fraction of them
with probability at least $1-\beta$. For fixed
$\alpha,\beta,\varepsilon,\delta$, a number of equations linear
in $d$ suffices. We give the precise bound and proof in
Appendix~\ref{sec:linear-equalities-corollary}.

\paragraph{Applying the structural mechanism.}
We first describe a private procedure for selecting a subspace from
a finite sequence $P=(u_i)_{i=1}^m$ of vectors in $\mathbb F^D$.
We will then apply
this procedure to the positive examples in a labeled sample.
Set the output universe to be
\[
\mathcal U=\{V\le\mathbb F^D:\dim V\le d\},
\]
the collection of all linear subspaces of dimension at most $d$.
For each input $P$, the candidate solutions are the distinct
subspaces in this universe that can be spanned by vectors from $P$:
\[
\operatorname{Out}(P)
=\{\operatorname{span}\{u_i:i\in I\}:I\subseteq[m],\;
\dim\operatorname{span}\{u_i:i\in I\}\le d\}.
\]
This set is finite and contains the zero subspace, obtained by
taking $I=\varnothing$. We score a candidate by the number of
positions whose vectors it contains, and rank it by its codimension:
\[
q_P(V)=|\{i\in[m]:u_i\in V\}|,
\qquad
\rho(V)=D-\dim V.
\]
Thus the score rewards covering many input vectors, while the rank
favors smaller subspaces. With privacy parameters $0<\varepsilon\le1$ and
$0<\delta<1$, and with $\Delta=B=1$ as verified below, these choices
specify Algorithm~\ref{alg:structural-em}: it samples
$V\in\operatorname{Out}(P)$ with probability proportional to
\[
\exp\!\left(\frac\varepsilon2[q_P(V)+\lambda(D-\dim V)]\right),
\qquad
\lambda=1+\frac2\varepsilon\log\frac8\delta.
\]

For the analysis, we define the maps between neighboring candidate
sets. Suppose $P$ is obtained from $Q$ by inserting one
vector $u$ at an arbitrary position. Define
\[
F_{P,Q}(V)=\operatorname{span}(Q|_V)
\quad\text{for }V\in\operatorname{Out}(P),
\qquad
F_{Q,P}(W)=W
\quad\text{for }W\in\operatorname{Out}(Q).
\]
Here $Q|_V$ is the subsequence of entries of $Q$ lying in $V$,
in their original order, and the span of a sequence means the
linear span of its entries.
The first map retains the part of $V$ spanned by the remaining
input vectors; its image belongs to $\operatorname{Out}(Q)$.
The reverse map is inclusion, since every candidate generated from
$Q$ is also generated from $P$.

These maps fix every common candidate. For the map
$F_{P,Q}$, we now reverse the insertion: delete the newly inserted
entry $u$ from $P$ to recover $Q$. If
$V\in\operatorname{Out}(P)\setminus\operatorname{Out}(Q)$, then every
subsequence of $P$ spanning $V$ must include this entry; otherwise,
$V$ would already belong to $\operatorname{Out}(Q)$. Hence
\[
V=\operatorname{span}(F_{P,Q}(V)\cup\{u\}),
\qquad u\notin F_{P,Q}(V).
\]
Thus $V$ loses exactly one dimension under the map, and its rank
increases by one.

The deletion map $F_{P,Q}$ is at most two-to-one. Each image $W$
has the fixed preimage $W$ itself, and at most one additional
preimage, namely $V=\operatorname{span}(W\cup\{u\})$. When this
additional preimage exists, its dimension is one larger than
that of $W$, so its rank is one smaller.

Recall that $B$ bounds the number of preimages at each positive
rank gap (Theorem~\ref{thm:structural-em}): at gap $t\ge1$, there must be at most $B^t$ preimages.
Here there is at most one preimage at gap one and none at larger
gaps. The fixed preimage has gap zero and does not enter this
count. Thus we may take $B=1$.

For a fixed subspace, adding or deleting one vector changes its
score by at most one. Also, $V$ and $F_{P,Q}(V)$ contain exactly
the same vectors of $Q$, so their scores on $P$ and $Q$ differ
by at most one. The reverse inclusion map satisfies the same
score bound. This verifies the sensitivity and score-preservation
conditions with $\Delta=1$.
Theorem~\ref{thm:structural-em} therefore gives
$(\varepsilon,\delta)$-privacy under addition/deletion on all input sequences.

To obtain the PAC learner, form $P$ by retaining the vectors of
the positively labeled entries of $S$, in their original order,
and apply this procedure with
privacy parameters $(\varepsilon,\delta)$.
Adding or deleting one labeled example changes $P$ by at most
one vector. The returned subspace indicator therefore inherits
$(\varepsilon,\delta)$-privacy.

On a realizable sample of size $n$, the positive examples lie in
the target subspace, which has dimension at most $d$. Every candidate
therefore lies in the target subspace, so there are no false positives:
missed positive examples are exactly the classification errors.
The full positive span is a candidate containing all the positive
examples. The rank range is at most $d$, and, for $n\ge d$,
$|\operatorname{Out}(P)|\le(en/d)^d$.
Theorem~\ref{thm:structural-em} and the VC generalization bound of Lemma~\ref{thm:vc-generalization} give
Theorem~\ref{thm:private-linear-subspaces}.
The absence of $d$ from the preimage bound $B=1$ is what avoids
the extra $\log d$ factor in the general Littlestone result.
In Section~\ref{sec:subspace-proof} we give a complete proof of
Theorem~\ref{thm:private-linear-subspaces}.

\subsection{Littlestone Classes: Ranks and Local Counting}
\label{sec:littlestone-overview}

We now describe the rank and the data-dependent candidate sets
that allow the structural exponential mechanism to yield our
main result. Fix a nonempty class $\mathcal H$ with
$d=\operatorname{LDim}(\mathcal H)\ge1$.
The rank is a combinatorial abstraction of the linear algebraic codimension,
defined for every class of finite Littlestone dimension.
We first develop the definition through the simple example of the class of singletons,
then explain its relation to subspace codimension and how it
guides the construction of candidates.

\paragraph{Model-theoretic origins.}
Littlestone dimension is the finite combinatorial form of Shelah's
$2$-rank~\cite{shelah1990classification,MalliarisMoran2022}.
In learning theory, it provides an abstraction of majority voting
that applies even to infinite classes. Given the class $\mathcal H$
and a query point $x$, compare
\[
\operatorname{LDim}(\{f\in\mathcal H:f(x)=0\})
\quad\text{and}\quad
\operatorname{LDim}(\{f\in\mathcal H:f(x)=1\}).
\]
The label whose class has larger Littlestone dimension plays the
role of the majority label, even when both classes are infinite.
This gives the Standard Optimal Algorithm
(SOA) predictor of Littlestone~\cite{littlestone87}, defined formally
in Definition~\ref{def:soa}.
Our rank relies on the Littlestone dimension of \emph{version spaces} (Definition~\ref{def:version-spaces}):
the subclasses of $\mathcal H$ consistent with a prescribed
collection of labeled examples. To construct the candidate set for
the structural exponential mechanism, we select, based on the sample,
a small collection of version spaces and use their SOA predictors
as candidates. Together, these candidates realize every labeling
of the sample induced by $\mathcal H$.
To bound the size of this candidate set, we also use ordinary
majority votes in finite
\emph{good families}: families of predictors in which, at each
point, almost all members assign the same label.
These families appear in the stable regularity framework of
Malliaris and Shelah~\cite{MalliarisShelah2014} and its
learning-theoretic development~\cite{MalliarisMoran2021}.
The treatment below is self-contained and requires no model theory.

\paragraph{Rank at least zero: local realizability.}
For a predictor $h:\mathcal X\to\{0,1\}$ and a finite set
$A\subseteq\mathcal X$, write
\[
\mathcal H[h,A]=\{f\in\mathcal H:f|_A=h|_A\}
\]
for the version space obtained by prescribing the labels of $h$
on $A$. For an integer $k\ge1$, we say that $h$ is
\emph{$k$-locally realizable} with respect to $\mathcal H$ if this
version space is nonempty for every $A$ of size at most $k$.
Thus, the labels of $h$ on any set of at most $k$ points are
consistent with $\mathcal H$: any finite collection of labeled points certifying that $h\notin\mathcal H$, if one exists, must contain at least $k+1$ points.

We now take $k=d+1$ and use the $(d+1)$-locally realizable
predictors as our output universe and as the predictors of
rank at least zero:
\[
F_0=\mathcal U
=\{h:\mathcal H[h,A]\ne\emptyset
\text{ for every }A\subseteq\mathcal X\text{ with }|A|\le d+1\}.
\]
Every member of $\mathcal H$ belongs to $F_0$, but $F_0$ can
strictly contain $\mathcal H$.

For example, let $\mathcal X$ be infinite and let
$\mathcal H=\{h_x:x\in\mathcal X\}$ be the singleton class,
where $h_x(z)=\mathbf1\{z=x\}$. This class has Littlestone
dimension $d=1$. Its locally realizable completion is
\[
F_0=\mathcal H\cup\{\mathbf0\},
\]
where $\mathbf0$ is the constant-zero predictor.
Indeed, the zero labeling on any two points is realized by a
singleton outside those points, while a predictor that labels
two distinct points positively is not locally realizable.

\paragraph{Higher ranks.}
To have rank at least one, a predictor must first belong to
$F_0$. We then require that prescribing its labels on any set
of at most $d$ points leaves a version space of Littlestone
dimension at least one. The predictors satisfying this
additional condition form $F_1$.

If $d\ge2$, rank at least two requires membership in $F_1$
and requires that every set of at most $d-1$ points leave a
version space of dimension at least two. These predictors form
$F_2$. We continue in this way: each step raises the required
remaining dimension by one and reduces the number of prescribed
labels by one. At the last step, membership in $F_d$ requires
membership in $F_{d-1}$ and dimension at least $d$ after prescribing
the label of any single point.

\begin{definition}[Rank]\label{def:rank}
Let $F_0=\mathcal U$. For $r=1,\ldots,d$, recursively define
\[
F_r=\{h\in F_{r-1}:
\operatorname{LDim}(\mathcal H[h,A])\ge r
\text{ whenever }|A|\le d+1-r\}.
\]
For $h\in\mathcal U$, set
$\rho(h)=\max\{r\in\{0,\ldots,d\}:h\in F_r\}$.
\end{definition}

Thus $F_r$ consists of the predictors of rank at least $r$.
Higher rank means that a large Littlestone dimension remains
after prescribing the predictor's labels on small sets.
The rank depends only on $\mathcal H$ and the predictor, and
is independent of the sample.

\paragraph{Basic properties of the rank.}
Three combinatorial properties will guide the construction:
predictors of higher rank form classes of smaller Littlestone
dimension (Lemma~\ref{lem:rank-dimension}); overwhelming majorities
increase rank (Lemma~\ref{lem:majority-rank}); and small sets of
labeled points distinguish predictors from others of equal or higher
rank (Lemma~\ref{lem:public-witness}). We state and prove these
properties together in Section~\ref{sec:rank-properties}.

For the singleton class, the constant-zero predictor has rank one:
prescribing a zero at any point leaves all the other singletons,
whose Littlestone dimension is still one. Each singleton $h_x$
has rank zero, because prescribing its positive label at $x$
leaves the singleton version space $\{h_x\}$, of dimension zero.

Although $F_0\supseteq\mathcal H$ and this inclusion may be strict, its Littlestone dimension is the same like $\mathcal H$:
\[
\operatorname{LDim}(F_0)=\operatorname{LDim}(\mathcal H)=d.
\]
Indeed, any path in a tree of depth $d+1$ uses at most $d+1$
points, so local realizability would transfer a shattered tree of this depth from $F_0$ to $\mathcal H$; see also
Lemma~\ref{lem:rank-dimension}.

\paragraph{Relation to codimension.}
For the subspace class $\mathcal H_{d,D}$ of
Section~\ref{sec:subspace-overview}, suppose that $D\ge d+1$.
Then the abstract rank of the indicator $h_V$ of a subspace $V$
is its codimension relative to the dimension cap:
\[
\rho(h_V)=d-\dim V.
\]
This measures how many dimensions remain available after
specifying $V$. Adding the constant $D-d$ recovers the geometric
rank $D-\dim V$ used for subspaces; this shift changes neither
rank gaps nor sampling probabilities.

\paragraph{Constructing the candidate set.}
Our goal is to construct a small data-dependent candidate set
that satisfies the requirements of the structural exponential
mechanism.
We use version spaces both in the original class $\mathcal H$
and in the enlarged class $\mathcal U$.

The small-witness property (Lemma~\ref{lem:public-witness})
gives each $h\in\mathcal U$ a witness set $A_h$ of at most
$d-\rho(h)$ points such that the
version space in $\mathcal U$ agreeing with $h$ on $A_h$
contains only $h$ and predictors of lower rank. We denote this
\emph{witness version space} by $Q_h$.
The corresponding version space in $\mathcal H$ has SOA predictor
$h$. We fix one such witness for each predictor, independently
of the sample.

\begin{samepage}
We now describe the construction of $\operatorname{Out}(S)$ for a given sample $S$, formalized in Algorithm~\ref{alg:candidate-sets}:

\begin{enumerate}
\item Partition $\mathcal U$ into the nonempty version spaces
induced by all labelings of the sample points.
\item Within each version space, proceed from rank $d$ down to
rank zero. At each rank, select every predictor that does \underline{not}
belong to the witness version space of a predictor selected at
a higher rank.
\item Take the union of the selected predictors over all these
version spaces to obtain the candidate set
$\operatorname{Out}(S)=\mathcal C(S)$.
\end{enumerate}

\end{samepage}
A useful observation (Lemma~\ref{lem:trace-coverage}) is that, by construction, every predictor is either selected or belongs to the witness version space of a selected predictor of strictly
higher rank in the same version space. In particular,
$\operatorname{Out}(S)$ realizes every labeling of the sample
induced by $\mathcal U$.

For subspaces with $D>d$, this yields
precisely the subspaces of dimension at most $d$ spanned by
subsets of the sample points, regardless of their labels.
The construction in Theorem~\ref{thm:private-linear-subspaces}
uses only positively labeled points, which
reduces the candidate set when many labels are zero.

\paragraph{The neighboring-sample map.}
For every ordered pair of neighboring samples $S,T$, where $T$
is obtained from $S$ by adding or deleting one example, define
$F_{S,T}:\operatorname{Out}(S)\to\operatorname{Out}(T)$ as follows.

\begin{enumerate}
\item If $g\in\operatorname{Out}(S)$ also belongs to
$\operatorname{Out}(T)$, set $F_{S,T}(g)=g$.
\item Otherwise, consider the version space in $\mathcal U$ of
predictors agreeing with $g$ on every point of $T$.
Since $g\notin\operatorname{Out}(T)$, Lemma~\ref{lem:trace-coverage} guarantees
that $g$ belongs to the witness version space~$Q_h$ of some
predictor $h$ selected at a strictly higher rank in this same
version space. Set $F_{S,T}(g)=h$, choosing one such $h$ by a fixed rule.
\end{enumerate}

Thus the map fixes common candidates, strictly increases the
rank of disappearing candidates, and preserves their labels
on the points of $T$.

{
\paragraph{Counting the candidates.}
We need to bound both the total number of candidates, for
utility, and the number that can map to a single candidate on
a neighboring sample, for privacy. 
The argument rests on two
combinatorial lemmas about good families.
Recall that a finite nonempty family $G$ is \emph{good} if, at every
point, fewer than a $1/[2(d+1)]$ fraction of its members disagree
with its pointwise majority predictor $b_G$.
Consequently, on any set of at most $d+1$ points, more than half
of $G$ agrees with $b_G$ simultaneously on all those points.

We summarize good-subfamily extraction and the majority property
below. Their full statements are Lemmas~\ref{lem:good-subset}
and~\ref{lem:majority-rank}, respectively.
\begin{enumerate}
\item \textbf{Finding a good subfamily.}
A finite family of Littlestone dimension at most $k$ with more
than $[2(d+1)]^k$ members contains a good subfamily with at least
two members.
\item \textbf{Majority increases rank.}
For $0\le s<d$, the majority predictor of a good family of at
least two distinct predictors, all of rank at least $s$, has
rank at least $s+1$.
\end{enumerate}

These lemmas imply that the selected candidates of any one rank
in a version space cannot contain a good family with
at least two members. To see this, fix one of these version spaces~$V$
and suppose that some selected predictors of rank $s<d$
form such a good family $G$. Its majority $b_G$ still belongs to $V$,
because all members of $G$ agree on the sample points.
By Lemma~\ref{lem:majority-rank}, $b_G$ has rank greater than $s$.
Thus, $b_G$ was considered before rank $s$ and lies in the
witness version space of a predictor selected at a rank above
$s$ (possibly $b_G$ itself) (Lemma~\ref{lem:trace-coverage}). Goodness ensures that some member
of $G$ agrees with $b_G$ on the at most $d$ witness points.
It therefore belongs to that same witness version space and
could not have been selected at rank $s$, a contradiction; this is the avoidance argument of Lemma~\ref{lem:majority-avoidance}.
At rank $d$, there is at most one predictor, since
$\operatorname{LDim}(F_d)\le0$ (Lemma~\ref{lem:rank-dimension}).

We now apply Lemma~\ref{lem:good-subset} to turn this exclusion into
counting bounds. The predictors of rank at least $s$ have
Littlestone dimension at most $d-s$ (Lemma~\ref{lem:rank-dimension}), so at most
$[2(d+1)]^{d-s}$ rank-$s$ candidates can be selected in any one
version space. Summing over ranks and using the
Sauer--Shelah bound (recalled in Section~\ref{sec:dimensions}) on the number of nonempty version spaces
induced by the $n$ sample points gives a total candidate count
whose logarithm is $O(d\log(en))$
for $n\ge d$.

For the local bound, fix a predictor $h$ of rank $r$.
Inside its witness version space $Q_h$, the predictors of rank
at least $s\le r$ have Littlestone dimension at most $r-s$
(Lemma~\ref{lem:cylinder-dimension}). The same argument therefore
bounds the number of selected rank-$s$ predictors in $Q_h$ and
a fixed version space by $[2(d+1)]^{r-s}$.
The exponent is the rank gap $r-s$, which is exactly the form
needed for the structural mechanism's preimage bound.
The precise counting statements appear in
Lemma~\ref{lem:candidate-counts}.

\paragraph{Applying the mechanism.}

Using the number of correctly classified examples as the score gives
$\Delta=1$. For a fixed image and rank gap $t\ge1$, its preimages
agree on the common sample points and hence occupy at most two
version spaces on $S$. The local count (Lemma~\ref{lem:candidate-counts}) therefore bounds their number
by $2[2(d+1)]^t\le[4(d+1)]^t$, giving $B=4(d+1)$.
The rank range is at most $d$, and every realizable sample has a
candidate making no mistakes (Lemma~\ref{lem:trace-coverage}). The structural mechanism's utility
bound (Equation~\eqref{eq:structural-em-utility}), the candidate count  (Lemma~\ref{lem:candidate-counts}), and VC generalization  (Lemma~\ref{thm:vc-generalization}) now yield
Theorem~\ref{thm:private-littlestone-pac}.

\section{Technical Preliminaries}
\label{sec:preliminaries}

We use natural logarithms, except in the iterated logarithm
$\log^*$. For $x>0$, $\log^*x$ is the least number of successive
applications of $\log_2$ needed to obtain a value at most $1$;
in particular, $\log^*x=0$ for $0<x\le1$.
A sample (or dataset) is a finite sequence $S=(z_1,\ldots,z_n)$,
with repeated entries allowed. We write $|S|=n$ for its length
and count every position in empirical losses. A subsequence
retains the entries at selected positions in their original order.
For an integer $n\ge0$, write $[n]=\{1,\ldots,n\}$, with
$[0]=\varnothing$.

\subsection{Learning and Privacy}

Let $\mathcal H\subseteq\{0,1\}^{\mathcal X}$ be a nonempty binary
concept class. For a distribution $\mathcal P$ on labeled examples,
write $L_{\mathcal P}(h)=\Pr_{(x,y)\sim\mathcal P}[h(x)\ne y]$.
The distribution is realizable by $\mathcal H$ if its labels are
given almost surely by some $h_*\in\mathcal H$.
An $(\alpha,\beta)$-PAC learner, on $n$ independent examples from
any such distribution, outputs a hypothesis with
$L_{\mathcal P}(h)\le\alpha$ with probability at least $1-\beta$.
The probability is over both the sample and the algorithm's
randomness. A learner is \emph{proper} if it always outputs a
member of $\mathcal H$.

We also use the realizable online model, in which examples arrive
sequentially and may be chosen adversarially, rather than being
drawn independently from a fixed distribution as in PAC learning.
At round $t$, the learner observes $x_t$, predicts
$\widehat y_t\in\{0,1\}$, and then receives the label~$y_t$.
The sequence must be consistent with some $h_*\in\mathcal H$,
meaning that $y_t=h_*(x_t)$ for every $t$.
Performance is measured by the number of mistakes
$\widehat y_t\ne y_t$; see Littlestone~\cite{littlestone87}.

Two samples are neighboring if one is obtained from the other
by inserting one example at any position or deleting one entry,
with the remaining entries in the same relative order. A randomized algorithm
$\mathcal A$ is $(\varepsilon,\delta)$-differentially private if,
for every ordered pair of neighboring
samples $S,T$ and every measurable output event $E$,
\[
\Pr[\mathcal A(S)\in E]
\le e^\varepsilon\Pr[\mathcal A(T)\in E]+\delta.
\]
Privacy is required on all inputs, independently of realizability.
Pure privacy is the case $\delta=0$.

\subsection[Dimensions, Version Spaces, and Generalization]{Dimensions, Version Spaces, and Generalization}\label{sec:dimensions}

Given points $x_1,\ldots,x_m$, a predictor $h$ \emph{realizes}
a binary labeling $(y_1,\ldots,y_m)\in\{0,1\}^m$ if
$h(x_i)=y_i$ for every $i\in[m]$. A class
$\mathcal G\subseteq\{0,1\}^{\mathcal X}$ realizes this labeling
if it contains such a predictor. A finite set of points is
\emph{shattered} by $\mathcal G$ if every binary labeling of it
is realized by $\mathcal G$. The VC dimension is the largest
size of a shattered set, or infinity if these sizes are unbounded.
We next define version spaces and the tree-based analogue of this dimension.

\begin{definition}[Version spaces]\label{def:version-spaces}
For a class $\mathcal G\subseteq\{0,1\}^{\mathcal X}$ and a
labeled sequence $T=((x_i,y_i))_{i=1}^m$, its \emph{version space}
consistent with $T$ is
\[
\mathcal G|_T
=\{h\in\mathcal G:h(x_i)=y_i\text{ for every }i\in[m]\}.
\]
We also call this the restriction of $\mathcal G$ to $T$.
In particular,
\[
\mathcal G|_{(x,y)}=\{h\in\mathcal G:h(x)=y\}.
\]
For a predictor $h:\mathcal X\to\{0,1\}$ and a finite set
$A\subseteq\mathcal X$, write
\[
\mathcal G[h,A]=\{g\in\mathcal G:g|_A=h|_A\}
\]
for the version space obtained by labeling $A$ according to $h$.
\end{definition}

\begin{definition}[Littlestone dimension]

A \emph{mistake tree} of depth $m$ over $\mathcal X$ is a complete
binary tree in which every root-to-leaf path has $m$ edges.
Each internal node is labeled by a point of $\mathcal X$.
Of the two edges leaving any internal node, one is labeled $0$
and the other is labeled $1$.
A class $\mathcal G\subseteq\{0,1\}^{\mathcal X}$ \emph{shatters}
the tree if, for every root-to-leaf path with successive node labels
$x_1,\ldots,x_m$ and edge labels $y_1,\ldots,y_m$, there exists
$h\in\mathcal G$ satisfying $h(x_t)=y_t$ for all $t\in[m]$.
The Littlestone dimension $\operatorname{LDim}(\mathcal G)$ is the
largest depth of a mistake tree shattered by $\mathcal G$,
or infinity if these depths are unbounded.
We set $\operatorname{LDim}(\emptyset)=-1$.
\end{definition}

Throughout this section, assume
$$
d=\operatorname{LDim}(\mathcal{H})<\infty.
$$

\begin{definition}[Empirical loss and realizability]
For a sample $S=((x_i,y_i))_{i=1}^n$ and a hypothesis
$h:\mathcal{X}\to\{0,1\}$, define
$$
\ell_S(h)
=
\bigl|\{i\in[n]:h(x_i)\ne y_i\}\bigr|,
\qquad
\operatorname{OPT}_S
=
\min_{h\in\mathcal{H}}\ell_S(h).
$$
Here $\ell_S(h)\in\{0,1,\ldots,n\}$ counts mistakes and is
integer-valued. In contrast, the normalized empirical loss is
$L_S(h)=\ell_S(h)/n$ for $n\ge1$.
The sample is \emph{realizable} by $\mathcal{H}$ if
$\operatorname{OPT}_S=0$, equivalently,
$\mathcal{H}|_S\ne\emptyset$.
\end{definition}

\begin{definition}[Standard Optimal Algorithm classifier]\label{def:soa}
For a nonempty subclass $\mathcal{G}\subseteq\mathcal{H}$,
define $\operatorname{SOA}_{\mathcal{G}}:
\mathcal{X}\to\{0,1\}$ by
$$
\operatorname{SOA}_{\mathcal{G}}(x)
=
\operatorname*{arg\,max}_{y\in\{0,1\}}
\operatorname{LDim}\bigl(\mathcal{G}|_{(x,y)}\bigr),
$$
with ties broken in favor of $0$.
\end{definition}

The SOA selects a label whose corresponding version space
has maximal Littlestone dimension. The version space for the
opposite label always has smaller dimension than $\mathcal G$:
$$
\operatorname{LDim}\left(
\mathcal{G}|_{(x,1-\operatorname{SOA}_{\mathcal{G}}(x))}
\right)
<
\operatorname{LDim}(\mathcal{G}).
$$

Every set of $k$ points shattered in the VC sense gives a shattered
Littlestone tree of depth $k$, so
$\operatorname{VCdim}(\mathcal G)\le\operatorname{LDim}(\mathcal G)$.
By the Sauer--Shelah bound~\cite{Sauer}, if
$\operatorname{VCdim}(\mathcal G)\le d$ and $n\ge d\ge1$, the
number of nonempty version spaces obtained by labeling
any $n$ points is at most
$\sum_{i=0}^d\binom ni\le(en/d)^d$.

\begin{lemma}[VC generalization {\cite{BEHW}}]
\label{thm:vc-generalization}
Let $\mathcal G$ be a binary class with
$\operatorname{VCdim}(\mathcal G)\le d$, where $d\ge1$.
There is a universal constant $C$ such that, if
\[
n\ge C\frac{d\log(e/\alpha)+\log(1/\beta)}{\alpha},
\]
then, with probability at least $1-\beta$ over an independent
sample from any distribution $\mathcal P$, every $h\in\mathcal G$
with $\ell_S(h)/n\le\alpha/2$ satisfies
$L_{\mathcal P}(h)\le\alpha$.
\end{lemma}

The following lemma is folklore; we include a proof for completeness.
\begin{lemma}[Dimensions of the subspace class]
\label{lem:subspace-dimensions}
For every field $\mathbb F$ and $1\le d\le D$,
$\operatorname{VCdim}(\mathcal H_{d,D})
=\operatorname{LDim}(\mathcal H_{d,D})=d$.
\end{lemma}
\begin{proof}
Any $d$ linearly independent vectors are shattered: each subset
is labeled positively by its span and the other basis vectors lie
outside that span. This proves the VC lower bound.
For the upper bound, consider the online rule that predicts $1$
exactly on the span of previously observed positive examples.
On a sequence labeled by $h_{V_*}\in\mathcal H_{d,D}$, it never
makes a false-positive mistake. Every false-negative mistake adds
a linearly independent vector to this span, so there are at most
$d$ mistakes. A shattered tree of depth $d+1$ would force any
deterministic online rule to make $d+1$ mistakes, a contradiction.
\end{proof}

\section{Proof of the Structural Mechanism}
\label{sec:mechanism-proof}

Write $S\sim T$ for the insertion/deletion relation on sequences
defined in Section~\ref{sec:preliminaries}. Assume the four hypotheses of
Theorem~\ref{thm:structural-em} for every ordered pair $(S,T)$
with $S\sim T$.

\begin{proof}[Proof of Theorem~\ref{thm:structural-em}]
We use the notation and parameter ranges of
Algorithm~\ref{alg:structural-em} and Theorem~\ref{thm:structural-em}.
All probabilities below are over the internal randomness of
$\mathcal A$.

We may assume $\Delta=1$: replace every $q_S$ by $q_S/\Delta$
and $\lambda$ by $\lambda/\Delta$.
This leaves the sampling distribution unchanged and makes both
score bounds at most one. We use $q_S$ and $\lambda$ for the
normalized quantities, so
$\lambda=1+(2/\varepsilon)\log(8B/\delta)$.
At the end, multiplying the utility bound by $\Delta$ recovers the
statement in the original score units.

We first prove privacy. Fix datasets $S\sim T$ and choose maps
$F_{S,T}:\operatorname{Out}(S)\to\operatorname{Out}(T)$ and
$F_{T,S}:\operatorname{Out}(T)\to\operatorname{Out}(S)$ satisfying
the theorem's hypotheses. Let
$I=\operatorname{Out}(S)\cap\operatorname{Out}(T)$ be their common
output set. The key step is to bound the probability of outputs that
disappear when the dataset changes from $S$ to $T$.
For $D\in\{S,T\}$ and $h\in\operatorname{Out}(D)$, define the weight
and normalizing constant by
\[
w_D(h)=\exp\!\left(\frac{\varepsilon}{2}
[q_D(h)+\lambda\rho(h)]\right),
\qquad Z_D=\sum_{h\in\operatorname{Out}(D)}w_D(h).
\]
Each $Z_D$ is finite and strictly positive because
$\operatorname{Out}(D)$ is finite and nonempty.
For every measurable event $E\subseteq\mathcal U$, write
\[
P_D(E)=\Pr[\mathcal A(D)\in E]
=\frac{1}{Z_D}\sum_{h\in E\cap\operatorname{Out}(D)}w_D(h).
\]

\paragraph{Weights of disappearing outputs.}
Fix $g\in\operatorname{Out}(S)\setminus I$, and put
$h=F_{S,T}(g)$ and $t=\rho(h)-\rho(g)$.
The rank-increase axiom makes $t$ a positive integer.
Score preservation under the map gives
\[
\frac{w_S(g)}{w_T(h)}
\le e^{\varepsilon(1-\lambda t)/2}
=e^{\varepsilon(1-t)/2}
\left(\frac{\delta}{8B}\right)^t
\le\left(\frac{\delta}{8B}\right)^t.
\]
For each fixed image $h$ and rank gap $t$, there are at most $B^t$
such preimages, each of weight at most
$w_T(h)(\delta/(8B))^t$.
Summing first over the preimages, then over the rank gaps and images,
gives
\begin{equation}
\begin{aligned}
\sum_{g\in\operatorname{Out}(S)\setminus I}w_S(g)
&\le\sum_{h\in\operatorname{Out}(T)}w_T(h)
\sum_{t\ge1}B^t\left(\frac{\delta}{8B}\right)^t\\
&=Z_T\frac{\delta}{8-\delta}
\le\frac{\delta}{4}Z_T.
\end{aligned}
\label{eq:structural-exclusive-weight}
\end{equation}
Applying the same argument to the reverse map $F_{T,S}$ gives
\eqref{eq:structural-exclusive-weight} with $S$ and $T$ interchanged.

\paragraph{Privacy.}
For $h\in I$, its rank is the same on both datasets, so low
sensitivity gives
$e^{-\varepsilon/2}\le w_S(h)/w_T(h)\le e^{\varepsilon/2}$.
Splitting each normalizing constant into common and disappearing
outputs therefore yields
\[
Z_T\le\left(e^{\varepsilon/2}+\frac{\delta}{4}\right)Z_S,
\qquad
Z_S\le\left(e^{\varepsilon/2}+\frac{\delta}{4}\right)Z_T.
\]
In particular, $Z_T/Z_S<2$, since $\varepsilon\le1$ and $\delta<1$.
Dividing \eqref{eq:structural-exclusive-weight} by $Z_S$ now shows
that the probability of a disappearing output is small:
\[
P_S(\operatorname{Out}(S)\setminus I)
\le\frac{\delta}{4}\frac{Z_T}{Z_S}
\le\frac{\delta}{2}.
\]
For an arbitrary measurable event $E\subseteq\mathcal U$, combining
this with the weight comparison for common outputs gives
\begin{align*}
P_S(E)
&\le e^{\varepsilon/2}\frac{Z_T}{Z_S}P_T(E\cap I)
+\frac{\delta}{2}\\
&\le e^\varepsilon P_T(E\cap I)
+\frac{e^{\varepsilon/2}\delta}{4}P_T(E\cap I)
+\frac{\delta}{2}\\
&\le e^\varepsilon P_T(E)+\delta,
\end{align*}
where the last step uses $e^{\varepsilon/2}<2$.
Interchanging $S,T$ proves the other privacy inequality.

\paragraph{Utility.}
Fix a dataset $S$ and a failure probability $0<\beta<1$, and let
$\widehat h=\mathcal A(S)$ be the random output. Write
$N=|\operatorname{Out}(S)|$ and define the regularized score
$r_S(h)=q_S(h)+\lambda\rho(h)$ for $h\in\operatorname{Out}(S)$, with
$r_* =\max_{h\in\operatorname{Out}(S)}r_S(h)$.
Recall that $R_S$ is the difference between the largest and smallest
ranks on $\operatorname{Out}(S)$ (Equation~\eqref{eq:rank-range}).
For every $a>0$, direct comparison with a maximizing candidate gives
\[
\Pr[r_S(\widehat h)<r_*-a]
\le N e^{-\varepsilon a/2}.
\]
Take $a=(2/\varepsilon)\log(N/\beta)$, and let $h_*$ maximize
$q_S(h)$ over $\operatorname{Out}(S)$.
On the resulting event of probability at least $1-\beta$,
\begin{align*}
q_S(\widehat h)
&\ge q_S(h_*)-\lambda(\rho(\widehat h)-\rho(h_*))-a\\
&\ge q_S(h_*)-\lambda R_S-a.
\end{align*}
Since $0<\varepsilon\le1$, the normalized rank coefficient satisfies
\[
\lambda=\frac2\varepsilon\left[
\frac\varepsilon2+\log\frac{8B}{\delta}\right]
\le\frac2\varepsilon\log\frac{16B}{\delta},
\]
where we used $\varepsilon/2\le\log 2$.
Substituting this bound and returning to the original score units
proves~\eqref{eq:structural-em-utility}.
\par
\end{proof}

\section{Application to Subspace PAC Learning}
\label{sec:subspace-proof}
\label{sec:linear-subspaces-algorithm}

We first analyze a private span-selection procedure using
Theorem~\ref{thm:structural-em}, and then turn it into a proper
PAC learner with the same privacy parameters. Throughout, the dimension cap $d$ and ambient space
$\mathbb F^D$ are fixed independently of the input.

\subsection{Private Span Selection}

For a finite sequence $P=(u_i)_{i=1}^m$ of vectors, write
\[
\mathcal C_d(P)
=\{\operatorname{span}\{u_i:i\in I\}:I\subseteq[m],\;
\dim\operatorname{span}\{u_i:i\in I\}\le d\}.
\]
This finite set is nonempty because it contains $\{0\}$.
We write $\operatorname{span}(P)$ for the span of its entries
and $P|_V$ for the subsequence of entries lying in $V$.
With privacy parameters $0<\varepsilon\le1$ and $0<\delta<1$, apply
Algorithm~\ref{alg:structural-em} with
\[
\operatorname{Out}(P)=\mathcal C_d(P),\quad
q_P(V)=|\{i\in[m]:u_i\in V\}|,\quad
\rho(V)=D-\dim V,\quad \Delta=B=1.
\]
Denote this procedure by $\operatorname{SpanSelect}_{d,\varepsilon,\delta}$.
It uses $\lambda=1+(2/\varepsilon)\log(8/\delta)$ and samples with
probability proportional to
$\exp((\varepsilon/2)[q_P(V)+\lambda(D-\dim V)])$.

\begin{lemma}[Span selection]
\label{thm:linear-subspaces-utility}
The procedure $\operatorname{SpanSelect}_{d,\varepsilon,\delta}$ is
$(\varepsilon,\delta)$-differentially private under addition/deletion
on all finite sequences of vectors. Its output $V$ always satisfies
$V\subseteq\operatorname{span}(P)$ and $\dim V\le d$.
If $m=|P|$ and $1\le r=\dim\operatorname{span}(P)\le d$, then,
{for every $0<\gamma<1$,} with probability at least $1-\gamma$,
\[
|\{i\in[m]:u_i\notin V\}|
\le\frac2\varepsilon\left[
r\log\frac{16}{\delta}
+r\log\frac{em}{r}+\log\frac1\gamma\right].
\]
If $\operatorname{span}(P)=\{0\}$, there are no missed vectors.
\end{lemma}

\begin{proof}
We verify the four axioms of Theorem~\ref{thm:structural-em}.
For a fixed $V$, adding or deleting one vector changes $q_P(V)$
by at most one, so the low-sensitivity axiom holds with $\Delta=1$.

Let $P$ be obtained from $Q$ by inserting one vector $u$, and define
$F_{P,Q}(V)=\operatorname{span}(Q|_V)$.
The image is generated by vectors of $Q$ and has dimension at most
$\dim V\le d$, so it belongs to $\mathcal C_d(Q)$.
The reverse map $F_{Q,P}$ is inclusion.
If $V\in\mathcal C_d(Q)$, then $Q|_V$ spans $V$, so the map
fixes every common output.

If $V\in\mathcal C_d(P)\setminus\mathcal C_d(Q)$, any subsequence
of $P$ spanning $V$ must include the inserted entry $u$. Hence
\[
V=\operatorname{span}(F_{P,Q}(V)\cup\{u\}),\qquad
u\notin F_{P,Q}(V).
\]
It follows that $\dim F_{P,Q}(V)=\dim V-1$ and
$\rho(F_{P,Q}(V))=\rho(V)+1$.
For each $W\in\mathcal C_d(Q)$, the subspace $W$ itself is a
preimage of the same rank. Any other preimage must equal
$\operatorname{span}(W\cup\{u\})$ and has rank $\rho(W)-1$.
Thus $W$ has at most one preimage whose rank is one less than its
own, and none whose rank is smaller by two or more.
This proves the preimage bound with $B=1$.
The reverse inclusion map fixes every input, so it has no
preimages of smaller rank.

Finally, $V$ and $F_{P,Q}(V)$ contain exactly the same vectors of $Q$.
Their scores on $P$ and $Q$ therefore differ by at most one.
For the inclusion map the same score bound is immediate.
All four axioms hold on arbitrary input sequences, proving privacy.

For utility, suppose $1\le r=\dim\operatorname{span}(P)\le d$.
The full span belongs to $\mathcal C_d(P)$ and has score $m$.
The rank range on the candidate set is $r$, and every candidate has
a basis of at most $r$ input vectors. Since $m\ge r$,
\[
|\mathcal C_d(P)|\le\sum_{j=0}^r\binom mj\le(em/r)^r.
\]
Substituting $\Delta=B=1$, $R_P=r$, and this candidate count
into~\eqref{eq:structural-em-utility} gives the claimed bound on
$m-q_P(V)$, the number of missed vectors.
When the input span is zero, the zero subspace is the only candidate
and contains every input vector.
\end{proof}

\subsection{From Span Selection to a Proper PAC Learner}

\begin{algorithm}[H]
\caption{Private PAC learning of subspace indicators}
\label{alg:private-linear-subspaces}
\begin{algorithmic}[1]
\Require Labeled sample $S=((x_i,y_i))_{i=1}^n$ in
$\mathbb F^D\times\{0,1\}$; dimension cap $d$;
privacy parameters $0<\varepsilon\le1$, $0<\delta<1$.
\State Form $P=(x_i)_{i\in[n]:y_i=1}$.
\State Sample
$V\gets\operatorname{SpanSelect}_{d,\varepsilon,\delta}(P)$.
\State \Return $h_V:x\mapsto\mathbf1\{x\in V\}$.
\end{algorithmic}
\end{algorithm}

\begin{proof}[Proof of Theorem~\ref{thm:private-linear-subspaces}]
Inserting or deleting one labeled example inserts or deletes
at most one vector in the sequence of positive examples.
Lemma~\ref{thm:linear-subspaces-utility}
therefore gives $(\varepsilon,\delta)$-privacy for the returned
subspace indicator by postprocessing.
%This argument holds even on unrealizable samples.
The dimension cap ensures that the output always belongs to
$\mathcal H_{d,D}$, so the learner is proper.

Now let $S$ be realizable by $h_{V_*}\in\mathcal H_{d,D}$.
Since every entry of $P$ lies in $V_*$, every candidate satisfies $V\subseteq V_*$.
Thus there are no false-positive errors, and
\[
\ell_S(h_V)=|P|-q_P(V).
\]
In particular, we use the score on positive examples only;
negative examples need not satisfy any structural score condition
on unrealizable inputs.
The full positive span is a candidate of zero loss.
For $n\ge d$, the rank range is at most $d$ and
$|\mathcal C_d(P)|\le\sum_{j=0}^d\binom nj\le(en/d)^d$.
Applying Theorem~\ref{thm:structural-em} with failure probability
$\beta/2$ gives
\[
\ell_S(h_V)\le\frac2\varepsilon\left[
d\log\frac{16}{\delta}
+d\log\frac{en}{d}+\log\frac2\beta\right].
\]
For
\[
n\ge C\frac{d\log\frac{e}{\alpha\varepsilon\delta}
+\log\frac1\beta}{\alpha\varepsilon},
\]
with a sufficiently large universal constant $C$, the right-hand
side is at most $\alpha n/2$.
The cancellation of $d$ in $\log(en/d)$ is what avoids a
$d\log d$ term in this sample bound.
Finally, Lemma~\ref{lem:subspace-dimensions} gives VC dimension $d$,
so Lemma~\ref{thm:vc-generalization}, with failure probability
$\beta/2$, transfers empirical error at most $\alpha/2$ to population
error at most $\alpha$. A union bound completes the proof.
\end{proof}

\section{Application to Littlestone Classes}
\label{sec:littlestone-proof}

We recall the construction from Section~\ref{sec:littlestone-overview}
and verify the four axioms of Theorem~\ref{thm:structural-em}.
Fix a nonempty class $\mathcal H$ of Littlestone dimension $d\ge1$.
{We first define the output universe and rank, then establish
some basic combinatorial properties in Section~\ref{sec:rank-properties}.
Section~\ref{sec:littlestone-counts} uses these properties to construct
and count the candidates sets associated with the input samples. Sections~\ref{sec:littlestone-axioms}
and~\ref{sec:littlestone-utility} then prove privacy and the PAC guarantee.
\par}

\paragraph{Universe and rank.}
We allow outputs outside $\mathcal H$, but require their labels
on every set of at most $d+1$ points to be consistent with
$\mathcal H$. This gives the output universe
\[
F_0=\mathcal U
=\{h\in\{0,1\}^{\mathcal X}:\mathcal H[h,A]\ne\emptyset
\text{ for every }A\subseteq\mathcal X\text{ with }|A|\le d+1\}.
\]
The rank (Definition~\ref{def:rank}) measures how much Littlestone dimension survives when
we prescribe a predictor's labels on small sets. For $r=1,\ldots,d$, set
\[
F_r=\{h\in F_{r-1}:\operatorname{LDim}(\mathcal H[h,A])\ge r
\text{ whenever }|A|\le d+1-r\}.
\]
For $h\in\mathcal U$, its rank is
\[
\rho(h)=\max\{r\in\{0,\ldots,d\}:h\in F_r\}.
\]

\subsection{Basic Properties of the Rank}
\label{sec:rank-properties}

Before analyzing the learning algorithm, we establish three basic
combinatorial properties of the rank. 
% Higher-rank predictors form
% classes of smaller Littlestone dimension; overwhelming majorities
% increase rank; and each predictor can be distinguished from all
% other predictors of equal or higher rank using a small set of labeled
% points. These properties concern the rank itself and will supply the
% dimension bounds and witnesses used in the candidate construction.

\begin{lemma}[Dimension decreases with rank]
\label{lem:rank-dimension}
For every integer $0\le s\le d$, the class $F_s$ of predictors of
rank at least $s$ has Littlestone dimension at most $d-s$.
In particular, $\operatorname{LDim}(\mathcal U)=d$.
\end{lemma}

\begin{lemma}[Overwhelming majorities increase rank]
\label{lem:majority-rank}
Let $0\le s<d$, and let $G$ be a finite family of at least two
distinct predictors in $\mathcal U$ of rank at least $s$.
If, at every point, fewer than a $1/[2(d+1)]$ fraction of $G$
disagree with its majority predictor $b_G$, then $b_G$ has rank
at least $s+1$.
\end{lemma}

\begin{lemma}[Small witnesses distinguish predictors]
\label{lem:public-witness}
For every predictor $h\in\mathcal U$ of rank $r$, there is a set
$A_h\subseteq\mathcal X$ of at most $d-r$ points such that $h$
is the only predictor of rank at least $r$ agreeing with $h$ on
$A_h$. The same set satisfies
\[
\operatorname{LDim}(\mathcal H[h,A_h])=r,\qquad
\operatorname{SOA}_{\mathcal H[h,A_h]}=h.
\]
Here $\operatorname{SOA}$ is the Standard Optimal Algorithm
predictor, defined in Definition~\ref{def:soa}.
\end{lemma}

\begin{proof}[Proof of Lemma~\ref{lem:rank-dimension}]
We first handle $s=0$, proving the stronger equality $\operatorname{LDim}(\mathcal U)=d$. Since $\mathcal H\subseteq\mathcal U$, we have
$\operatorname{LDim}(\mathcal U)\ge d$.
For the reverse inequality, suppose that $\mathcal U$ shatters a complete binary tree of depth
$d+1$. Fix any root-to-leaf path, and let $f\in\mathcal U$ realize
its labels. The set $A$ of points queried along this path has
size at most $d+1$. By the definition of $\mathcal U$, some
$h\in\mathcal H[f,A]$ realizes the same path.
Thus $\mathcal H$ also shatters this tree, contradicting
$\operatorname{LDim}(\mathcal H)=d$.
Therefore $\operatorname{LDim}(\mathcal U)=d$.

For $1\le s\le d$, suppose towards contradiction that $F_s$ shatters a tree of depth $d-s+1$.
For each root-to-leaf path, let $A$ be its set of points and choose
$g\in F_s$ realizing its labels. Since $|A|\le d+1-s$, the
version space $\mathcal H[g,A]$ has dimension at least~$s$.
Append a tree of depth $s$ shattered by this version space below
the corresponding leaf. Doing this for every leaf gives a tree
of depth $d+1$ shattered by $\mathcal H$, a contradiction.
\end{proof}

\begin{proof}[Proof of Lemma~\ref{lem:majority-rank}]
Write $b=b_G$. The assumed pointwise agreement and a union bound imply that, for every
set $A$ of at most $d+1$ points,
\[
|\{g\in G:g|_A=b|_A\}|>|G|/2.
\]
Since $|G|\ge2$, at least two members agree with $b$ on all of $A$.
First, $b\in\mathcal U$: for each such $A$, take an agreeing
$g\in G\subseteq\mathcal U$, giving
$\mathcal H[b,A]=\mathcal H[g,A]\ne\emptyset$.
For $1\le j\le s$ and $|A|\le d+1-j$, the same choice gives
\[
\operatorname{LDim}(\mathcal H[b,A])
=\operatorname{LDim}(\mathcal H[g,A])\ge j,
\]
since $g\in F_s\subseteq F_j$. Hence $b\in F_s$.

Now fix $|A|\le d-s$. Choose distinct $f,g\in G$ agreeing
with $b$ on $A$, and a point $x$ where they differ.
Necessarily $x\notin A$. The two branches of $\mathcal H[b,A]$
at $x$ are $\mathcal H[f,A\cup\{x\}]$ and
$\mathcal H[g,A\cup\{x\}]$. Both have dimension at least $s$,
because $|A\cup\{x\}|\le d+1-s$ and $f,g\in F_s$.
For $s=0$, use membership in $\mathcal U$.
Putting $x$ above shattered trees of depth $s$ in the two
branches proves $\operatorname{LDim}(\mathcal H[b,A])\ge s+1$.
This holds for every such $A$, so $b\in F_{s+1}$.
\end{proof}

\begin{proof}[Proof of Lemma~\ref{lem:public-witness}]
Let $r=\rho(h)$. We first choose $A_h$ with the claimed size and
dimension. If $r=d$, take $A_h=\emptyset$, so
$\mathcal H[h,A_h]=\mathcal H$ has dimension $d$.
For $0\le r<d$, the fact that $h\notin F_{r+1}$ supplies $A_h$
with $|A_h|\le d-r$ and dimension less than $r+1$, while
$h\in F_r$ gives dimension at least $r$. Thus in every case
$|A_h|\le d-r$ and $\operatorname{LDim}(\mathcal H[h,A_h])=r$.

We next show that $\operatorname{SOA}_{\mathcal H[h,A_h]}=h$.
For any $x\in\mathcal X$, we have $|A_h\cup\{x\}|\le d+1-r$,
and hence
\[
r\le\operatorname{LDim}(\mathcal H[h,A_h\cup\{x\}])
\le\operatorname{LDim}(\mathcal H[h,A_h])=r.
\]
The first inequality follows from $h\in F_r$ when $r\ge1$
and from local realizability when $r=0$.
Thus the branch corresponding to $h(x)$ has dimension $r$.
The opposite branch has smaller dimension: otherwise, putting
$x$ above two shattered trees of depth $r$ would give dimension
at least $r+1$ for $\mathcal H[h,A_h]$.
Therefore the SOA predicts $h(x)$ at every point $x$.

To prove the distinguishing property, let $g\in\mathcal U$ have
rank at least $r$ and agree with $h$ on $A_h$.
Then $\mathcal H[g,A_h]=\mathcal H[h,A_h]$ has dimension $r$.
The same branch argument applies to $g$: membership in $F_r$
(or local realizability if $r=0$) shows that its label at every
point is the SOA label of this version space. Hence
$g=\operatorname{SOA}_{\mathcal H[h,A_h]}=h$.
\end{proof}

\subsection{Candidate Construction and Counting}
\label{sec:littlestone-counts}

In this subsection, we present Algorithm~\ref{alg:candidate-sets}
for constructing the candidate set of the structural exponential
mechanism and prove bounds on the number of candidates, both in
total and inside each witness version space. The small-witness
lemma (Lemma~\ref{lem:public-witness}) provides the exclusion rule
used in the construction, while the dimension bounds
(Lemmas~\ref{lem:rank-dimension} and~\ref{lem:cylinder-dimension})
and the majority property (Lemma~\ref{lem:majority-rank}) control
how many candidates survive this rule.
\paragraph{Witness version spaces.}
For each $h\in\mathcal U$, fix one witness $A_h$ supplied by
Lemma~\ref{lem:public-witness}.
The witnesses are fixed once and for all,
independently of the input sample $S$, and the same witnesses are
used to construct the candidate set for every sample.
Define
\[
Q_h=\mathcal U[h,A_h]
=\{g\in\mathcal U:g|_{A_h}=h|_{A_h}\}.
\]
The distinguishing property in Lemma~\ref{lem:public-witness}
ensures that every member of $Q_h$
other than $h$ has strictly smaller rank. These witness version
spaces guide candidate selection.

\paragraph{Selecting candidates.}
We now define the candidate set for each input sample.

For a sample $S=((x_i,y_i))_{i=1}^n$ and a labeling
$u=(u_1,\ldots,u_n)\in\{0,1\}^n$, write
\[
V_u(S)=\mathcal U|_{((x_i,u_i))_{i=1}^n}
=\{h\in\mathcal U:h(x_i)=u_i\text{ for all }i\in[n]\}.
\]
The nonempty version spaces $V_u(S)$ partition $\mathcal U$;
we consider all labelings $u$, regardless of the observed labels $y_i$.

Algorithm~\ref{alg:candidate-sets} constructs $\mathcal C(S)$ by
processing each version space from higher to lower rank.
{At each rank, we keep every predictor that does not
belong to the witness version space of a candidate already selected
at a higher rank in the same version space.}

\begin{algorithm}[H]
\caption{Candidate set construction for Littlestone classes}
\label{alg:candidate-sets}
\begin{algorithmic}[1]
\Require Sample $S=((x_i,y_i))_{i=1}^n$; universe $\mathcal U$,
rank $\rho$, and witness version spaces $(Q_h)_{h\in\mathcal U}$ as above.
\ForAll{nonempty version spaces $V=V_u(S)$, $u\in\{0,1\}^n$}
    \State $\mathcal D(V)\gets\emptyset$.
    \For{$s=d,d-1,\ldots,0$}
        \State $\displaystyle T_s(V)\gets
        \{h\in V:\rho(h)=s,\;h\notin\bigcup_{g\in\mathcal D(V)}Q_g\}$.
        \State $\mathcal D(V)\gets\mathcal D(V)\cup T_s(V)$.
    \EndFor
\EndFor
\State $\mathcal C(S)\gets\bigcup_{V}\mathcal D(V)$,
where the union is over the version spaces processed above.
\State \Return $\mathcal C(S)$.
\end{algorithmic}
\end{algorithm}
We set
$\operatorname{Out}(S)=\mathcal C(S)$.
Lemma~\ref{lem:trace-coverage} proves $\operatorname{Out}(S)\neq\emptyset$, and
Lemma~\ref{lem:candidate-counts} establishes an upper bound on its size.

\paragraph{Counting candidates.}
We will bound the total number of candidates and, within each
version space $V_u(S)$, the number of candidates of a given rank
belonging to a fixed witness version space~$Q_h$ (Lemma~\ref{lem:candidate-counts}). The latter bound will control
how many candidates can map to a fixed output in the privacy
analysis. To this end, we first complement the dimension bound
for $F_s$ in Lemma~\ref{lem:rank-dimension} with a bound on the
dimension of $Q_h\cap F_s$.

\begin{lemma}[Dimension inside a witness version space]
\label{lem:cylinder-dimension}
Let $h\in\mathcal U$ have rank $r$. For every $0\le s\le r$,
the class of predictors in $Q_h$ whose rank is at least $s$ has
Littlestone dimension at most $r-s$; that is,
$\operatorname{LDim}(Q_h\cap F_s)\le r-s$.
In particular, $Q_h\cap F_r=\{h\}$.
\end{lemma}

\begin{proof}
Suppose, towards a contradiction, that $Q_h\cap F_s$
shatters a tree of depth $r-s+1$.
For each root-to-leaf path, let $A$ be its set of points and
choose $g\in Q_h\cap F_s$ realizing its labels. Since
\[
|A_h\cup A|\le(d-r)+(r-s+1)=d+1-s,
\]
membership in $F_s$ (Definition~\ref{def:rank}) gives
$\operatorname{LDim}(\mathcal H[g,A_h\cup A])\ge s$ when $s\ge1$.
When $s=0$, local realizability of $g\in\mathcal U$ and
$|A_h\cup A|\le d+1$ ensure that $\mathcal H[g,A_h\cup A]$ is
nonempty, so its Littlestone dimension is at least $0=s$.
Every member of this version space agrees with $h$ on $A_h$
and realizes the path. Appending a shattered tree of depth $s$
below each leaf therefore gives a tree of depth $r+1$ shattered
by $\mathcal H[h,A_h]$, contradicting its dimension $r$.

Finally, taking $s=r$ gives $\operatorname{LDim}(Q_h\cap F_r)\le0$.
This intersection contains $h$ and cannot contain a second distinct
predictor, since two distinct predictors shatter a tree of depth one.
Hence $Q_h\cap F_r=\{h\}$.
\end{proof}

To bound the number of selected predictors of rank $s$, we combine
the dimension bounds for $F_s$ and $Q_h\cap F_s$
(Lemmas~\ref{lem:rank-dimension} and~\ref{lem:cylinder-dimension})
with the following extraction lemma.
For $0<\theta\le1/2$, a finite nonempty family $G$ is \emph{$\theta$-good} if, for
every $x\in\mathcal X$, one label is assigned by fewer than
$\theta|G|$ members of $G$. The other label defines its majority
predictor $b_G(x)$. In particular, fewer than $\theta|G|$ members
disagree with $b_G$ at any one point.
We call $G$ \emph{good} if it is $1/[2(d+1)]$-good.

The next lemma is the good subset extraction argument of
Malliaris and Shelah~\cite[Section~5]{MalliarisShelah2014},
stated in terms of Littlestone dimension. We include its short proof
for completeness.

\begin{lemma}[Good subset]\label{lem:good-subset}
Every finite nonempty family $G$ with
$\operatorname{LDim}(G)\le t$ contains a good subset
of size at least $|G|/[2(d+1)]^t$.
\end{lemma}
\begin{proof}
Write $\theta=1/[2(d+1)]$.
If the current family is not good, some point splits it into
two parts, each of size at least a $\theta$ fraction of the
family. At least one part has strictly smaller Littlestone
dimension. Keep that part and repeat. After at most $t$ steps
the remaining nonempty family is good, and its size is at
least~$\theta^t|G|$.
\end{proof}

Lemma~\ref{lem:majority-rank} shows that the majority of a good
family of at least two predictors of rank at least $s<d$ has rank
at least $s+1$. For the selection argument, we also need that
this majority avoids any witness version space avoided by all
members of the family.

\begin{lemma}[Majorities preserve avoidance of witness version spaces]
\label{lem:majority-avoidance}
Let $G\subseteq\mathcal U$ be a finite nonempty good
family, and let $h\in\mathcal U$. If every member of $G$ lies
outside $Q_h$, then its majority predictor $b_G$ also lies
outside $Q_h$.
%perhaps "disagrees with h on its witness set A_h" is better and more informative than "lies outside Q_h"? Maybe the contra-position is even better: if the majority vote agrees with h then all do... not sure...
\end{lemma}

\begin{proof}
Suppose that $b_G\in Q_h$. By a union bound, at most
$|A_h||G|/[2(d+1)]\le d|G|/[2(d+1)]<|G|$ members
of $G$ disagree with $b_G$ somewhere on $A_h$.
Thus some $g\in G$ agrees with $b_G$, and hence with $h$, on
$A_h$. This gives $g\in Q_h$, a contradiction.
\end{proof}

Before combining these facts, we record that every omitted
predictor belongs to the witness version space of a selected
predictor of strictly higher rank.

\begin{lemma}[Coverage]\label{lem:trace-coverage}
Let $S$ be a sample and $V=V_u(S)$ a nonempty version space.
For every $h\in V$, some $g\in\mathcal D(V)$ satisfies
$h\in Q_g$ and $\rho(g)\ge\rho(h)$.
If $h\notin\mathcal D(V)$, then $\rho(g)>\rho(h)$.
\end{lemma}

\begin{proof}[Proof of Lemma~\ref{lem:trace-coverage}]
If $h\in\mathcal D(V)$, take $g=h$.
Otherwise, the selection rule in Algorithm~\ref{alg:candidate-sets}
excludes~$h$ at rank $\rho(h)$ precisely because it lies in the
witness version space of a predictor selected at a strictly higher rank.
\end{proof}

Combining good-subfamily extraction (Lemma~\ref{lem:good-subset}),
rank increase (Lemma~\ref{lem:majority-rank}), avoidance of witness
version spaces (Lemma~\ref{lem:majority-avoidance}), and coverage
(Lemma~\ref{lem:trace-coverage}) gives the candidate counts for
Algorithm~\ref{alg:candidate-sets} below;
in the local bound, the exponent is the difference in ranks.
%This paragraph is somewhat dense; can be shorter and just say that we are now ready to state the main bounds or something. Maybe also remind what T_S(V) is 

\begin{lemma}[Global and local candidate counts]
\label{lem:candidate-counts}
For every sample $S$, every nonempty version space $V=V_u(S)$,
and every $0\le s\le d$,
\[
|T_s(V)|\le[2(d+1)]^{d-s}.
\]
For every $h\in\mathcal U$ of rank $r\ge s$,
\[
|T_s(V)\cap Q_h|\le[2(d+1)]^{r-s}.
\]
Consequently, for a sample of size $n\ge d$,
\[
|\mathcal C(S)|\le2[2(d+1)]^d(en/d)^d.
\]
\end{lemma}

\begin{proof}[Proof of Lemma~\ref{lem:candidate-counts}]
We first show that $T_s(V)$ contains no finite good
family of size at least two. For $s=d$, this follows from
$\operatorname{LDim}(F_d)\le0$ (Lemma~\ref{lem:rank-dimension}). For $s<d$, suppose such a family $G$ exists.
%suppose towards contradiction
Its majority $b_G$ has rank greater than $s$
by Lemma~\ref{lem:majority-rank}, and belongs to the same version space $V$. All members of $G$ avoid every witness version space of a predictor selected in
$V$ at a rank above $s$; Lemma~\ref{lem:majority-avoidance} shows that $b_G$ also
avoids these witness version spaces. But Lemma~\ref{lem:trace-coverage} puts
$b_G$ in the witness version space of a selected predictor of rank at least
$\rho(b_G)>s$, a contradiction.

If $|T_s(V)|>[2(d+1)]^{d-s}$, choose a finite subset $G_0$
larger than this bound. It has Littlestone dimension at most
$d-s$, since $G_0\subseteq F_s$ and Lemma~\ref{lem:rank-dimension} bounds the dimension of $F_s$.
Lemma~\ref{lem:good-subset} supplies a good subset of size
at least $|G_0|/[2(d+1)]^{d-s}>1$, contradicting the preceding
paragraph. This proves the global bound.

For the local bound, if $|T_s(V)\cap Q_h|>[2(d+1)]^{r-s}$,
choose a finite subset $G_0$ larger than this bound.
Lemma~\ref{lem:cylinder-dimension} gives
$\operatorname{LDim}(G_0)\le r-s$.
The good-subset lemma (Lemma~\ref{lem:good-subset}) again provides a good family of size
greater than one inside $T_s(V)$, a contradiction.

Finally,
\[
|\mathcal D(V)|\le\sum_{s=0}^d[2(d+1)]^{d-s}
\le2[2(d+1)]^d.
\]
There are at most $(en/d)^d$ nonempty version spaces $V_u(S)$
when $S$ contains $n\ge d$ points, by $\operatorname{VCdim}(\mathcal U)\le d$ (Lemma~\ref{lem:rank-dimension}) and the
Sauer--Shelah bound (recalled in Section~\ref{sec:dimensions}). Multiplication proves the total count.
\end{proof}

\subsection{Verifying the Structural Axioms}
\label{sec:littlestone-axioms}

We now construct the maps $F_{S,T}$ required by
Theorem~\ref{thm:structural-em}, using coverage
(Lemma~\ref{lem:trace-coverage}) to choose their images and the
local candidate bound (Lemma~\ref{lem:candidate-counts}) to
control their preimages.

Set $\operatorname{Out}(S)=\mathcal C(S)$ and
$q_S(h)=|S|-\ell_S(h)$ for $h\in\mathcal U$.
The candidate set is finite by Lemma~\ref{lem:candidate-counts}
and nonempty by coverage (Lemma~\ref{lem:trace-coverage}). The score of a fixed predictor changes
by at most one when an example is added or deleted.

Fix an ordered pair of neighboring datasets $(S,T)$, and let $R$
be the shorter sequence. One of $S,T$ equals $R$, and the other
is obtained from $R$ by inserting one example at some position.
For $g\in\mathcal C(S)\cap\mathcal C(T)$,
set $F_{S,T}(g)=g$. For any other $g\in\mathcal C(S)$, let $V_T$
be the version space in $\mathcal U$ consisting of predictors
that agree with $g$ on every point of $T$. Coverage (Lemma~\ref{lem:trace-coverage}) provides a selected predictor
$h\in\mathcal D(V_T)$ with $g\in Q_h$ and
$\rho(h)>\rho(g)$; set $F_{S,T}(g)=h$, choosing one such image
by a fixed rule. This defines a map to $\mathcal C(T)$ with the
required rank increase.

Because $g$ and $h$ agree on every sample point in $T$, they agree
on $R$. The contributions of $R$ to their scores are equal, and
the single additional example contributes either zero or one.
Thus
\[
|q_S(g)-q_T(F_{S,T}(g))|\le1.
\]

To bound preimages, fix $h\in\mathcal C(T)$ of rank $r$ and
a gap $t\ge1$. If $r-t<0$, there are no such preimages.
Otherwise, every preimage $g$ of rank $r-t$ belongs to $Q_h$
and agrees with $h$ on $R$. Since $S$ is either $R$ itself or
$R$ augmented by one example, these predictors occupy at most
two version spaces on $S$.
Inside each such version space $V$, every preimage belongs to
$T_{r-t}(V)\cap Q_h$. The local candidate bound (Lemma~\ref{lem:candidate-counts}) therefore gives
\[
|\{g\in F_{S,T}^{-1}(h):\rho(g)=r-t\}|
\le2[2(d+1)]^t\le[4(d+1)]^t.
\]
The construction works for every ordered pair of neighboring
datasets, covering both addition and deletion.
All four axioms hold with $\Delta=1$ and $B=4(d+1)$.

\subsection{Utility and PAC Learning}
\label{sec:littlestone-utility}

We finish by combining the mechanism's utility guarantee
(Equation~\eqref{eq:structural-em-utility}) with the candidate count
(Lemma~\ref{lem:candidate-counts}) to bound empirical loss, then
applying VC generalization (Lemma~\ref{thm:vc-generalization}) to
obtain the PAC guarantee.

Our learner first constructs $\mathcal C(S)$ using
Algorithm~\ref{alg:candidate-sets}, then applies
Algorithm~\ref{alg:structural-em} with this candidate set and the
preceding score and rank.

\begin{proof}[Proof of Theorem~\ref{thm:private-littlestone-pac}]
By Section~\ref{sec:littlestone-axioms}, all four axioms of
Theorem~\ref{thm:structural-em} hold with $\Delta=1$ and
$B=4(d+1)$. The theorem therefore gives
$(\varepsilon,\delta)$-privacy on all samples.

Let $S$ now be realizable by $h_*\in\mathcal H\subseteq\mathcal U$.
Writing $S=((x_i,y_i))_{i=1}^n$, recall that
$\ell_S(h)=\sum_{i=1}^n\mathbf 1_{\{h(x_i)\ne y_i\}}=n-q_S(h)$
counts the mistakes of $h$ on $S$.
Coverage (Lemma~\ref{lem:trace-coverage}) in the version space $\mathcal U|_S$ supplies a selected predictor
agreeing with $h_*$ on all sample points, hence of score $n$.
Thus $\max_{h\in\mathcal C(S)}q_S(h)=n$.
Rearranging Equation~\eqref{eq:structural-em-utility}
therefore gives an upper bound on
$\ell_S(\widehat h)=n-q_S(\widehat h)$, the number of mistakes
made by the output.
We apply that equation with failure probability $\beta/2$,
using $\Delta=1$, $B=4(d+1)$, $R_S\le d$, and, for $n\ge d$,
$|\operatorname{Out}(S)|=|\mathcal C(S)|\le2[2(d+1)]^d(en/d)^d$
from Lemma~\ref{lem:candidate-counts}.
With probability at least $1-\beta/2$, this gives

\[
\ell_S(\widehat h)\le\frac2\varepsilon\left[
d\log\frac{64(d+1)}{\delta}
+d\log\frac{2(d+1)en}{d}+\log\frac4\beta\right].
\]

For a sufficiently large universal $C$, this is at most
$\alpha n/2$ whenever
\[
n\ge C\frac{d\log\frac{ed}{\alpha\varepsilon\delta}
+\log\frac1\beta}{\alpha\varepsilon}.
\]
By Lemma~\ref{lem:rank-dimension},
$\operatorname{VCdim}(\mathcal U)\le d$.
Lemma~\ref{thm:vc-generalization}, with failure probability
$\beta/2$, now gives population error at most $\alpha$.
A union bound proves the claimed PAC guarantee.
\end{proof}

\begin{samepage}
\section*{Acknowledgements}
We are especially grateful to Maryanthe Malliaris for inspiring
and encouraging us to explore model-theoretic tools, most notably
ranks, in the study of private learning.

\paragraph{AI disclosure.}
The authors developed the strategy of studying linear subspaces
and extending the resulting linear-algebraic arguments to
Littlestone classes using model-theoretic ideas, and formulated
the structural exponential mechanism. The algorithms and proofs
were derived through interaction with ChatGPT Pro. ChatGPT Astra
also assisted with the writing. All use of these tools was guided
by the authors.
\par
\end{samepage}

\bibliographystyle{abbrv}
\bibliography{working-draft-refs}

\begin{thebibliography}{10}

\bibitem{AlonBMS20}
N.~Alon, A.~Beimel, S.~Moran, and U.~Stemmer.
\newblock Closure properties for private classification and online prediction.
\newblock In {\em {COLT}}, volume 125 of {\em Proceedings of Machine Learning Research}, pages 119--152. {PMLR}, 2020.

\bibitem{ALBMM22}
N.~Alon, M.~Bun, R.~Livni, M.~Malliaris, and S.~Moran.
\newblock Private and online learnability are equivalent.
\newblock {\em Journal of the ACM}, 69(4):28:1--28:34, 2022.

\bibitem{AlonLMM19}
N.~Alon, R.~Livni, M.~Malliaris, and S.~Moran.
\newblock Private {PAC} learning implies finite littlestone dimension.
\newblock In M.~Charikar and E.~Cohen, editors, {\em Proceedings of the 51st Annual {ACM} {SIGACT} Symposium on Theory of Computing, {STOC} 2019, Phoenix, AZ, USA, June 23-26, 2019}, pages 852--860. {ACM}, 2019.

\bibitem{BNS13}
A.~Beimel, K.~Nissim, and U.~Stemmer.
\newblock Characterizing the sample complexity of private learners.
\newblock In {\em ITCS}, pages 97--110. ACM, 2013.

\bibitem{BEHW}
A.~Blumer, A.~Ehrenfeucht, D.~Haussler, and M.~K. Warmuth.
\newblock Learnability and the vapnik-chervonenkis dimension.
\newblock {\em J. ACM}, 36(4):929--965, 1989.

\bibitem{bun2020equivalence}
M.~Bun, R.~Livni, and S.~Moran.
\newblock An equivalence between private classification and online prediction.
\newblock In {\em 2020 IEEE 61st Annual Symposium on Foundations of Computer Science (FOCS)}, pages 389--402. IEEE, 2020.

\bibitem{DiNi03}
I.~Dinur and K.~Nissim.
\newblock Revealing information while preserving privacy.
\newblock In {\em PODS}, pages 202--210. ACM, 2003.

\bibitem{DKMMN06}
C.~Dwork, K.~Kenthapadi, F.~McSherry, I.~Mironov, and M.~Naor.
\newblock Our data, ourselves: Privacy via distributed noise generation.
\newblock In S.~Vaudenay, editor, {\em EUROCRYPT}, volume 4004 of {\em Lecture Notes in Computer Science}, pages 486--503. Springer, 2006.

\bibitem{DMNS06}
C.~Dwork, F.~McSherry, K.~Nissim, and A.~Smith.
\newblock Calibrating noise to sensitivity in private data analysis.
\newblock In {\em TCC}, pages 265--284, 2006.

\bibitem{DR14}
C.~Dwork and A.~Roth.
\newblock The algorithmic foundations of differential privacy.
\newblock {\em Foundations and Trends in Theoretical Computer Science}, 9(3-4):211--407, 2014.

\bibitem{EHKV}
A.~Ehrenfeucht, D.~Haussler, M.~J. Kearns, and L.~G. Valiant.
\newblock A general lower bound on the number of examples needed for learning.
\newblock {\em Inf. Comput.}, 82(3):247--261, 1989.

\bibitem{ghazi2021sample}
B.~Ghazi, N.~Golowich, R.~Kumar, and P.~Manurangsi.
\newblock Sample-efficient proper pac learning with approximate differential privacy.
\newblock In {\em Proceedings of the 53rd Annual ACM SIGACT Symposium on Theory of Computing}, pages 183--196, 2021.

\bibitem{Hanneke16}
S.~Hanneke.
\newblock The optimal sample complexity of pac learning.
\newblock {\em J. Mach. Learn. Res.}, 17(1):1319--1333, Jan. 2016.

\bibitem{KMMST25}
H.~Kaplan, Y.~Mansour, S.~Moran, U.~Stemmer, and N.~Tur.
\newblock On differentially private linear algebra.
\newblock In {\em Proceedings of the 57th Annual ACM Symposium on Theory of Computing}, STOC '25, pages 2362--2373, New York, NY, USA, 2025. Association for Computing Machinery.

\bibitem{KLNRS08}
S.~P. Kasiviswanathan, H.~K. Lee, K.~Nissim, S.~Raskhodnikova, and A.~Smith.
\newblock What can we learn privately?
\newblock {\em SIAM J. Comput.}, 40(3):793--826, 2011.

\bibitem{Kearns94}
M.~J. Kearns and U.~V. Vazirani.
\newblock {\em An Introduction to Computational Learning Theory}.
\newblock MIT press, Cambridge, Massachusetts, 1994.

\bibitem{LWY25}
B.~Li, W.~Wang, and P.~Ye.
\newblock Private realizable-to-agnostic transformation with near-optimal sample complexity.
\newblock In {\em Proceedings of the Thirty Eighth Conference on Learning Theory}, volume 291 of {\em Proceedings of Machine Learning Research}, pages 3700--3722. PMLR, 2025.

\bibitem{littlestone87}
N.~Littlestone.
\newblock Learning quickly when irrelevant attributes abound: A new linear-threshold algorithm.
\newblock In {\em 28th Annual Symposium on Foundations of Computer Science (sfcs 1987)}, pages 68--77, 1987.

\bibitem{lyu2026private}
X.~Lyu.
\newblock Private learning of littlestone classes, revisited.
\newblock In {\em Proceedings of the 58th Annual ACM Symposium on Theory of Computing}, pages 2218--2229, 2026.

\bibitem{MalliarisMoran2021}
M.~Malliaris and S.~Moran.
\newblock Agnostic online learning and excellent sets.
\newblock arXiv:2108.05569, 2021.

\bibitem{MalliarisMoran2022}
M.~Malliaris and S.~Moran.
\newblock The unstable formula theorem revisited via algorithms.
\newblock arXiv:2212.05050, 2022.

\bibitem{MalliarisShelah2014}
M.~Malliaris and S.~Shelah.
\newblock Regularity lemmas for stable graphs.
\newblock {\em Transactions of the American Mathematical Society}, 366(3):1551--1585, 2014.

\bibitem{MT07}
F.~McSherry and K.~Talwar.
\newblock Mechanism design via differential privacy.
\newblock In {\em FOCS}, pages 94--103, 2007.

\bibitem{NST26}
K.~Nissim, U.~Stemmer, and E.~Tsfadia.
\newblock Invited open problem: Does differential privacy make {PAC} learning much harder?
\newblock In S.~Hanneke and T.~Lattimore, editors, {\em Proceedings of Thirty Ninth Conference on Learning Theory}, volume 336 of {\em Proceedings of Machine Learning Research}, pages 7129--7135. PMLR, 2026.

\bibitem{Sauer}
N.~Sauer.
\newblock On the density of families of sets.
\newblock {\em Journal of Combinatorial Theory, Series A}, 13(1):145 -- 147, 1972.

\bibitem{shelah1990classification}
S.~Shelah.
\newblock {\em Classification theory: and the number of non-isomorphic models}.
\newblock Elsevier, 1990.

\bibitem{Talagrand94}
M.~Talagrand.
\newblock Sharper bounds for {Gaussian} and empirical processes.
\newblock {\em The Annals of Probability}, 22(1):28--76, 1994.

\bibitem{VC}
V.~N. Vapnik and A.~Y. Chervonenkis.
\newblock On the uniform convergence of relative frequencies of events to their probabilities.
\newblock {\em Theory of Probability and its Applications}, 16(2):264--280, 1971.

\end{thebibliography}

\appendix
\section{A Private Solver for Linear Equalities}
\label{sec:linear-equalities-corollary}

The span-selection guarantee (Lemma~\ref{thm:linear-subspaces-utility}) yields a private solver through
the augmented-vector reduction used by Kaplan et al.~\cite{KMMST25}.

\begin{corollary}
Fix a field $\mathbb F$ and $d\ge1$. There is an
$(\varepsilon,\delta)$-differentially private algorithm on systems
of linear equations in $d$ variables, under addition/deletion of
one equation, with the following guarantee. If a system of $n$ equations
is consistent and
\[
n\ge C\frac{d\log\frac{e}{\alpha\varepsilon\delta}
+\log\frac1\beta}{\alpha\varepsilon},
\]
then, with probability at least $1-\beta$, the algorithm returns
a vector satisfying at least $(1-\alpha)n$ of the equations.
Here $C$ is a universal constant, and privacy holds on inconsistent
systems as well.
\end{corollary}

\begin{proof}
Write the input sequence of equations as $(a_i^\top x=b_i)_{i=1}^n$
and form the sequence $P=((a_i,-b_i))_{i=1}^n$ in $\mathbb F^{d+1}$.
Apply $\operatorname{SpanSelect}_{d,\varepsilon,\delta}$ to $P$,
and let $W$ be the output. Choose, by a fixed rule, a solution to
\[
w^\top(x,1)=0\qquad\text{for every }w\in W,
\]
if one exists, and return a fixed vector otherwise.
This last operation depends only on $W$; it can equivalently be
performed on any basis of $W$. Adding or deleting one equation
adds or deletes one augmented row. Span selection Lemma~\ref{thm:linear-subspaces-utility}) therefore gives
$(\varepsilon,\delta)$-privacy, and postprocessing preserves it
even on inconsistent inputs.

If the system has a solution $x_*$, then every augmented row lies
in the kernel of the nonzero linear functional
$w\mapsto w^\top(x_*,1)$. Thus
$\dim\operatorname{span}(P)\le d$.
Moreover, $W\subseteq\operatorname{span}(P)$, so the displayed
system always has a solution. Every augmented row contained in
$W$ corresponds to an equation satisfied by the returned vector.
Lemma~\ref{thm:linear-subspaces-utility} bounds the number of rows
outside $W$ by
\[
\frac2\varepsilon\left[
d\log\frac{16}{\delta}
+d\log\frac{en}{d}+\log\frac1\beta\right]
\]
with probability at least $1-\beta$. The asserted sample bound
makes this at most $\alpha n$.
\end{proof}

\section{Extension to Agnostic Learning}
\label{sec:agnostic-extension}

We explain how to apply the realizable-to-agnostic reduction
of~\cite[Theorem~3.6]{LWY25} under our addition/deletion
privacy convention.

\paragraph{Quantitative bound.}
Let $\mathcal H$ be a nonempty binary concept class, write
$v=\operatorname{VCdim}(\mathcal H)$, and fix $0<\delta_0<1$.
Let $m$ be the sample complexity of an
$(\varepsilon/2,\delta_0/3)$-private $(\alpha,\beta)$-PAC learner
for~$\mathcal H$ in the realizable setting.
By Footnote~\ref{fn:replacement-privacy}, this learner is
$(\varepsilon,\delta_0)$-private under replacement.
The reduction uses
\[
n=O\!\left(m+
\frac{v\log(1/\alpha)+\log(1/\beta)}{\alpha\varepsilon}
+\frac{v+\log(1/\beta)}{\alpha^2}\right)
\]
examples and is $(\varepsilon,\delta_0)$-private under replacement,
with excess error $O(\alpha)$ and failure probability
$O(\beta+n\delta_0)$. It also preserves properness.

To obtain excess error at most $\alpha$ and failure probability
at most $\beta$, apply the realizable-to-agnostic reduction with
sufficiently small constant multiples of $\alpha$ and $\beta$,
and take $\delta_0\le\min\{\delta,c\beta/n\}$ for a sufficiently
small absolute constant $c>0$. This absorbs the constants hidden
in the bounds above and controls the additional failure term
$O(n\delta_0)$. The resulting learner is $(\varepsilon,\delta)$-private
under replacement and has the desired learning guarantee on
i.i.d. samples of size $n$, where $n$ denotes the sample size
after these parameter choices.

\paragraph{Extending fixed-size learners.}
We now show how to convert the replacement-privacy guarantee
obtained above into addition/deletion privacy, the convention
used throughout this paper. To do so, we extend the agnostic
learner from samples of a fixed size $n$ to arbitrary input sizes.
Here $n$ is chosen in advance according to the stated sample
complexity bound, and we use the same fixed $n$ on every input,
including neighboring inputs.

Fix any labeled example $z_0$, independently of the input; its
label is arbitrary. For an input sequence of length $m<n$, append
$n-m$ copies of $z_0$. For $m>n$, choose a uniformly random
$n$-element subset of $[m]$ and retain those entries in their
original order. For $m=n$, retain the entire sequence.
In each case, apply a uniformly random permutation to the resulting
$n$ positions before running the learner. This last step can be
omitted if the learner is invariant under permutations.

For neighboring inputs, the padding and subsampling can be coupled
so that the retained entries admit a matching of size at least
$n-1$ with identical values. To see this for subsampling, assign
independent continuous random priorities to input positions, use
the same priorities for corresponding positions in the two inputs,
and retain the $n$ smallest priorities. Inserting one entry changes
the retained positions by at most one exchange. For padding,
match all common input entries and all but at most one padding copy.
Couple the final permutations by placing matched entries in the
same uniformly random positions; the two resulting sequences then
differ in at most one coordinate. Averaging the replacement privacy
inequality over this coupling proves
$(\varepsilon,\delta_0)$-privacy on inputs of every size.

The utility guarantee applies when at least $n$ genuine examples
are available. On an input of size $n$, no padding or subsampling
occurs. On a larger i.i.d. input, the uniformly selected $n$
examples are themselves i.i.d., and random ordering preserves
this distribution. Thus the extension preserves the agnostic
learning guarantee. Below size $n$, padding may affect the loss
and the predictions; we assert privacy but no accuracy guarantee
on these inputs. Properness is preserved on all inputs.

\end{document}